\documentclass{article}

\usepackage{microtype}
\usepackage{graphicx}
\usepackage{subfigure}
\usepackage{booktabs} % for professional tables

\usepackage{hyperref}

\usepackage{amsmath,amsfonts,bm}

\def\eqref#1{equation~\ref{#1}}
\def\1{\bm{1}}
\newcommand{\train}{\mathcal{D}}

\def\vtheta{{\bm{\theta}}}

\def\vf{{\bm{f}}}

\def\vx{{\bm{x}}}
\def\vy{{\bm{y}}}

\DeclareMathAlphabet{\mathsfit}{\encodingdefault}{\sfdefault}{m}{sl}
\SetMathAlphabet{\mathsfit}{bold}{\encodingdefault}{\sfdefault}{bx}{n}

\usepackage{xurl}

\usepackage[accepted]{icml2025}

\usepackage{amsmath}
\usepackage{amssymb}
\usepackage{mathtools}
\usepackage{amsthm}

\usepackage[capitalize,noabbrev]{cleveref}

\theoremstyle{plain}

\theoremstyle{definition}

\theoremstyle{remark}

\usepackage[textsize=tiny]{todonotes}

\icmltitlerunning{Beyond Scaling Curves: Internal Dynamics of Neural Networks Through the NTK Lens}

\begin{document}

\twocolumn[
\icmltitle{Beyond Scaling Curves: Internal Dynamics of Neural Networks Through the NTK Lens}



\icmlsetsymbol{equal}{*}

\begin{icmlauthorlist}
\icmlauthor{Konstantin Nikolaou}{yyy}
\icmlauthor{Sven Krippendorf}{xxx}
\icmlauthor{Samuel Tovey}{yyy}
\icmlauthor{Christian Holm}{yyy}
\end{icmlauthorlist}

\icmlaffiliation{yyy}{Institute of Computational Physics, University of Stuttgart, Stuttgart, Germany}
\icmlaffiliation{xxx}{Cavendish Laboratory and DAMTP, University of Cambridge, Cambridge, United Kingdom, CB3 0WA}


\icmlcorrespondingauthor{Konstantin Nikolaou}{knikolaou@icp.uni-stuttgart.de}

\icmlkeywords{Neural Networks, Neural Network Theory, Deep Learning, Deep Learning Theory, Scaling Laws, Neural Scaling Laws, Neural Tangent Kernel, Machine Learning, ICML}

\vskip 0.3in
]



\printAffiliationsAndNotice{}  % leave blank if no need to mention equal contribution

\begin{abstract}
    Scaling laws offer valuable insights into the relationship between neural network performance and computational cost, yet their underlying mechanisms remain poorly understood. 
    In this work, we empirically analyze how neural networks behave under data and model scaling through the lens of the neural tangent kernel (NTK). 
    This analysis establishes a link between performance scaling and the internal dynamics of neural networks.
    Our findings of standard vision tasks show that similar performance scaling exponents can occur even though the internal model dynamics show opposite behavior. 
    This demonstrates that performance scaling alone is insufficient for understanding the underlying mechanisms of neural networks.
    We also address a previously unresolved issue in neural scaling: how convergence to the infinite-width limit affects scaling behavior in finite-width models.
    To this end, we investigate how feature learning is lost as the model width increases and quantify the transition between kernel-driven and feature-driven scaling regimes.
    We identify the maximum model width that supports feature learning, which, in our setups, we find to be more than ten times smaller than typical large language model widths.
    % This work provides a foundation for quantifying the impact effects underlying neural scaling in empirical settings. 
\end{abstract}

\section{Introduction} %\looseness=-1

% Why do we care about scaling Laws
Neural scaling laws have emerged as empirical relationships that describe how the performance metrics of neural network (NN) models scale with key quantities such as model size, dataset size, and computational cost~\citep{KaplanScalingLawsNeural2020,henighanScalingLawsAutoregressive2020,hoffmannEmpiricalAnalysisComputeoptimal2022}. 
These laws relate computational investments to measures of model quality, such as the loss, and provide a practical framework for predicting the resources required to improve performance.
Understanding the origin and limitations shaping these laws may be one of the most important questions in current artificial intelligence (AI) research~\cite{hutterLearningCurveTheory2021}.
% “Chinchilla scaling” refines previous neural scaling laws by showing that model and dataset size should scale in tandem for optimal performance (Hoffmann et al., 2022). 

There are theories that offer conceptual insights into why scaling occurs~\citep{hutterLearningCurveTheory2021, sharma22a,bahriExplainingNeuralScaling2024b}.
To achieve this, a toy model is constructed that captures idealized scaling behavior, however, beyond the ideal setting, it is unclear how the predictions translate to real-world experiments and scaling phenomena observed in practice.
% they often fall short of explaining the scaling behaviour observed in practice. 
Rather than relying on a toy model for scaling, we aim to bridge the gap between theory and practice from the empirical side.
We study the inner workings of neural networks to investigate the mechanisms that influence and shape scaling predictions. 
This work establishes a baseline by using multi-layer architectures and simple vision benchmark datasets.

We connect loss scaling to network dynamics through the empirical Neural Tangent Kernel (NTK). 
We quantify the dynamics with two properties, the trace and the effective rank, representing the scale and dimensionality of the NTK, respectively.
Both quantities are conceptually motivated by statistical physics, where macroscopic observables encapsulate the behaviour of many interacting degrees of freedom. 
In that spirit we interpret the NTK trace and effective rank as collective variables: coarse descriptors that compress the high-dimensional learning dynamics into two interpretable numbers.
These physics-inspired variables, therefore, provide a coarse yet informative lens through which we examine how scaling transformations reshape a network's internal dynamics.
Using this analysis, we address the origin and limitations of neural scaling laws by investigating two key questions:
\begin{itemize}
    \item   \textbf{How does scaling model size and dataset size intrinsically affect a model’s behavior?}
            We find that even though scaling model and data size exhibits similar performance scaling exponents, the intrinsic model effects underlying this behavior are opposite:
            Model scaling affects the initial trace and reduces the effective rank of the NTK. In contrast, data scaling, affects the dynamics of the trace and reduces the effective rank of the NTK.
    \item   \textbf{At what model width can we identify the influence of the kernel regime?}
            When scaling model width to infinity, the training behaviour is known to be described by a static kernel, called the neural tangent kernel~\citep{jacotNeuralTangentKernel2018}. 
            In this limit, models lose feature learning capabilities~\citep{yangFeatureLearningInfiniteWidth2022}. 
            We empirically identify the model width at which this effect becomes dominant and show that, for our setups, this limit can be far below typical large language model (LLM) widths.
\end{itemize}

\section{Related Work}

\subsection{Neural Scaling Laws}

Neural scaling laws describe the quantitative trade-offs between model performance and 
investment in terms of data, model size, and compute resources~\citep{rosenfeldConstructivePredictionGeneralization2019, KaplanScalingLawsNeural2020,hoffmannEmpiricalAnalysisComputeoptimal2022}. 
These relationships are mostly expressed as power laws, with several works attempting 
to explain their origins~\citep{hutterLearningCurveTheory2021, sharma22a, bahriExplainingNeuralScaling2024b}. 
% However, varying scaling exponents and pre-factors observed across studies suggest that the explanations need to be refined.
However, it is unclear how these theoretical models translate to empirical settings like in LLMs~\citep{sharma22a}.
In this context, especially the role of learning dynamics remains under-explored; for example, \citet{hoffmannEmpiricalAnalysisComputeoptimal2022} attribute the discrepancies between their results and those of \citet{KaplanScalingLawsNeural2020} to differences in the learning-rate schedule, a key component of the dynamics.
The contribution of learning dynamics to scaling remains opaque in the current literature, and clarifying this relationship is the central aim of our work. \\
Particularly relevant is the work of \citet{bahriExplainingNeuralScaling2024b}, 
which distinguishes between two scaling regimes:
The resolution-limited regime, where model and data size scaling are constrained by 
the model's ability to resolve the data manifold, linking to~\citet{sharma22a}, and
the variance-limited regime, where scaling is constrained by the predicted variance due to
a dominating information bottleneck (such as little data or small models) 
while non-bottlenecked factors are scaled. 
% The latter regime can be thought of as learning to compensate for the bottleneck.
Besides allowing for the theoretical foundations analyzed in~\citet{bahriExplainingNeuralScaling2024b}, the variance-limited regime enables targeted investigations of the scaling behavior of applied settings and is the focus of our work.

\subsection{Neural Tangent Kernel Theory} %\looseness=-1

NTK theory characterizes the convergence behavior of infinitely wide NNs~\citep{jacotNeuralTangentKernel2018}.
% Various studies have examined the gap between finite and infinite networks, focusing on the empirical differences~\citep{ortiz-leeFiniteInfiniteNeural2020}, 
Several works have extended this approach: 
\citet{huangDynamicsDeepNeural2020} rewrote NN dynamics into a hierarchical set of differential equations, to cover the full learning dynamics,
\citet{yangFeatureLearningInfiniteWidth2022} investigating how to parametrize a network to avoid losing feature learning abilities in the infinite limit, and~\citep{meiMeanfieldTheoryTwolayers2019,geigerDisentanglingFeatureLazy2020} studied the differences of the NTK to the mean-field limit, where features in the asymptotic limit continue to evolve.\\
In the realm of finite models, research has started to focus on specific aspects of the NTK dynamics to understand more about the learning process. 
One line of research highlights the alignment of the NTK with task-relevant 
directions~\citep{baratinImplicitRegularizationNeural2021,atanasovNeuralNetworksKernel2021,shanTheoryNeuralTangent2022}. 
Among these works, \citet{ortiz-jimenezWhatCanLinearized2021} identified kernel rotation as an 
alignment mechanism.
Additionally, \citet{fortDeepLearningKernel2020} studied learning dynamics, revealing two learning phases: an initial evolution 
phase followed by minimal change of the NTK. 
\citet{toveyCollectiveVariablesNeural2024} also discovered a two-stage phenomenon using an NTK decomposition approach, which has further been applied to assess training data quality \citep{toveyPhenomenologicalUnderstandingNeural2023}.\\
% Works addressing the effect of kernel-like behavior on finite models 
Another line of work has focussed on the comparison between finite and infinite networks. 
\citet{chizatLazyTrainingDifferentiable2019b} investigated whether lazy training of linearized networks can explain the success of deep learning.
Most follow-up works focussed on disentangling the differences between finite and infinite networks, to identify in which cases one performs better than the other~\citep{leeFiniteInfiniteNeural2020,ortiz-jimenezWhatCanLinearized2021,seleznovaAnalyzingFiniteNeural2022}.
Yet, these works do not address, the transition between feature and kernel learning, which remains less explored. 
It is unclear how kernel-like behavior affects learning dynamics and scaling behavior in finite models and to what extent it influences scaling in state-of-the-art applications.

\section{Preliminaries}
\subsection{Decomposition of Learning Dynamics} \label{sec:ntk_decomposition}
%\looseness=-1

In this section, we briefly review the NTK definition from \citet{jacotNeuralTangentKernel2018} 
and introduce a decomposition into effective quantities to analyze learning dynamics.

Let $f(\vx, \vtheta)$ be the prediction of an NN\footnote{
    To keep the notation simple, we consider scalar-valued network functions here 
    and extend to the generic case of vector-valued functions in \cref{sec:ntk_decomposition_app}.
} with parameters $\vtheta$ and input $\vx$.
The network is trained on a dataset $\train = \{(\vx_i, y_i)\}_{i=1}^d = (\bm{X}, Y)$ with $d$ samples 
using a loss function $\mathcal{L}(f(\bm{X}, \vtheta), y) = \frac{1}{d} \sum_{i=1}^d \ell(f(\vx_i, \vtheta), y_i)$
and a gradient descent update rule. 
% the update rule
% \begin{equation}f
%     \vtheta_{t+1} = \vtheta_t - \eta \nabla_{\vtheta} \mathcal{L}(f(\bm{X}, \vtheta), \vy)
% \end{equation}
% where $\eta$ is the learning rate.
For asymptotically small step sizes, the update converges to gradient flow. 
% \begin{equation}
%     \frac{d \vtheta}{dt} = - \nabla_{\vtheta} \mathcal{L}(f(\bm{X}, \vtheta), \vy)
% \end{equation}
In this regime, the model is updated with continuous time $t$ and the evolution of the predictions
\begin{equation}
    \dot{f} = \frac{d f}{dt} 
    = - \nabla_{\vtheta} f(\bm{X}, \vtheta) \frac{d \vtheta}{dt}
    = - \bm{\Theta} \, \nabla_{f} \mathcal{L}(f(\bm{X}, \vtheta), \vy) , 
\end{equation}
is governed by the NTK 
$\bm{\Theta} = \nabla_{\vtheta} f(\bm{X}, \vtheta)^\top \nabla_{\vtheta} f(\bm{X}, \vtheta)$, 
describing the complexity and non-linear adaptation of the network function. 
In contrast to infinitely wide networks, the NTK of finite-width networks changes during training~\citep{jacotNeuralTangentKernel2018}.
% With basic linear algebra, we can diagonalize the NTK and split off the magnitude part in the form of the trace:
With basic linear algebra, we can split off the magnitude part in the form of the trace %$\bm{\Theta} =\hat{\bm{\Theta}} \, \text{Tr} \bm{\Theta} $, 
and diagonalize the remainder:
\begin{equation}
    \bm{\Theta} 
    = \text{Tr} (\bm{\Theta}) \,\bm{Q}^\top \bm{\Lambda} \bm{Q} 
    = \text{Tr} (\bm{\Theta}) \, \sum_{i=1}^d \lambda_i \bm{q}_i \bm{q}_i^\top
    \label{eq:ntk_decomposition}
\end{equation}
As the NTK is symmetric and positive semi-definite~\citep{jacotNeuralTangentKernel2018}, $\bm{Q}$ is the unitary matrix of eigenvectors $\bm{q}_i$
and $\bm{\Lambda}$ the diagonal matrix of normalized eigenvalues $\lambda_i$. \\
Extending on previous works \citep{kopitkovNeuralSpectrumAlignment2020a, baratinImplicitRegularizationNeural2021,krippendorfDualityConnectingNeural2022},
viewing the network evolution from the eigenbasis of the NTK, learning evolves through the orthonormal eigenmodes $\bm{q}_i$.
Each eigenmode is scaled by its corresponding eigenvalue and the magnitude $\lambda_i  \text{Tr} \bm{\Theta}$, describing the contribution of the mode to the learning process. \\
In the context of the NTK, it was proposed by~\citet{toveyPhenomenologicalUnderstandingNeural2023} to analyze the von Neumann entropy ${S^\text{vN}(M) = - \text{Tr} (M \ln M)}$ of the kernel.
For capturing the distribution of the eigenvalues,
% we extend this idea by introducing the effective rank~\citep{royEffectiveRankMeasure2007}
we equivalently consider the effective rank~\citep{royEffectiveRankMeasure2007} of the NTK, providing an intuitive interpretation of the entropy:% in the context of dimensionality measures:
\begin{equation}
    \Gamma(\bm{\Theta}) = \exp{S^\text{vN} \left( \bm{\Theta} \right)}    \label{eq:effective_rank}
\end{equation}
% This approach is further extended by incorporating the effective rank as a complementary measure (see Appendix~\ref{sec:effective_rank_entropy_app} for further details).
It quantifies the number of dominant eigenmodes and can thus be seen as the effective dimension of learning. \\
% In order to use the NTK trace and effective rank as a means to understand the learning and scaling of NNs, \cref{sec:mechanistic_interpretation_ntk_quantities} provides a mechanistic interpretation of these quantities. 
To summarize, we have split the NN learning dynamics into two parts:
The magnitude of the NTK is quantified by the trace $\text{Tr}(\bm{\Theta})$ and the eigendistribution of the NTK is quantified by its effective rank $\Gamma(\bm{\Theta})$.
Adopting a statistical-physics perspective, we treat $\text{Tr}(\bm{\Theta})$ and $\Gamma(\bm{\Theta})$ as collective variables, quantities that coarse-grain the combined influence of data samples and network parameters into global descriptors of the learning dynamics.
These quantities allow us to analyze aspects of the learning dynamics that are not accessible through the loss function alone, which is why we will refer to them as intrinsic quantities. 
We will use them as a lense to analyze how scaling of model and data size influences the learning dynamics of NNs. 

% \subsection{Computing Test NTK Quantities} %\looseness=-1
% Test data NTK
Given a data distribution, we can think of sampling train and test data as draws from this distribution.
Measuring a test loss then corresponds to measuring how well a model generalizes to the data distribution.
In this work, we extend this idea by measuring the NTK quantities $\text{Tr}(\bm{\Theta})$ and $\Gamma(\bm{\Theta})$ on the test data,
as we want to understand the dynamics of learning with respect to the data distribution 
that defines the underlying task.

% \newpage

%%%%%%%%%%%%%%%%%%%%%%%%%%%%%%%%%%%%%%%%%%%%%%%%%%%%%%%%%%%%%%%%%%%%%%%%%%%%%%%%%%%%%%%%%%%%%%%
% New Structure 
\newpage
\section{Scaling Effects} \label{sec:scaling_mechanisms} %\looseness=-1
% \subsection{Isolating Scaling Mechanisms} 
%\looseness=-1
% Goal of this section
% In this section, we analyze and compare the internal mechanisms of model and data scaling regimes. 
% Describe the experimental setup
% During the training of neural networks, the complexity of the utilized datasets is not homogeneous \citep{smithInstanceLevelAnalysis2014, zhangInstanceRegularizationDiscriminative2022}. 
% Certain aspects of the data may be more complex than others, requiring varying capacities for learning. 
% As a result, analyzing the training dynamics of such models could reveal both overparameterized and underparameterized aspects within the same training process. 
% To disentangle these dynamic contributions to scaling behavior, we investigate two key regimes:
% scaling large models while keeping data fixed and scaling large datasets while keeping model size fixed. 
% This approach corresponds to the variance-limited regime \citep{bahriExplainingNeuralScaling2024b}, where generalization is constrained primarily by either model capacity or data availability.\\
% In this section we analyze and compare how model scaling and data scaling influence the internal workings of neural networks.
% In this section, we will compare the intrinsic effects on learning dynamics when scaling model size to scaling data size. 
In this section, we compare how scaling model size versus scaling dataset size influences learning dynamics.
Because datasets often contain varying levels of complexity \citep{smithInstanceLevelAnalysis2014, zhangInstanceRegularizationDiscriminative2022}, some aspects of the data may be more difficult to learn than others. 
This heterogeneity in data complexity may lead to both over- and under-parameterized learning within the same training process, making it very hard to disentangle effects. 
To understand how these dynamic factors contribute to scaling behavior, we investigate two key regimes: 
1) increasing the model size while keeping a small dataset fixed, and 
2) increasing the dataset size while keeping a small model fixed.
This approach allows us to isolate the effects of model capacity and data availability on generalization in the over- and under parametrized ragime. 
It further corresponds to the variance-limited regime described in \citep{bahriExplainingNeuralScaling2024b}.\\
% where generalization is constrained primarily by either model capacity or data availability.\\
% During NN training, multiple factors can affect both the training process 
% and the model's ability to generalize. 
% To gain a clearer understanding of neural scaling, we study scaling contributions within 
% the \textit{variance limited regime}~\citep{bahriExplainingNeuralScaling2024b}. 
% This regime refers to a situation where one factor, or "bottleneck", plays a dominant role in limiting performance. 
% By adding a governing bottleneck, we can reduce the number of independent variables, 
% allowing us to better isolate and analyze the underlying behavior\footnote{
%     For a more detailed explanation of the isolation of scaling mechanisms, 
%     please refer to Appendix~\ref{sec:experimental_methods}.
% }.
% We will investigate two cases: 
% Training on small datasets while increasing already large models, 
% and training small models on increasingly large datasets.\\
Building upon the foundation laid in~\citet{bahriExplainingNeuralScaling2024b}, we analyze NNs trained on standard datasets, such as MNIST, Fashion-MNIST, and CIFAR-10, using cross-entropy loss and the Adam optimizer~\citep{kingmaAdamMethodStochastic2017b}. 
For all models, we use LeCun initialization~\citep{lecunEfficientBackProp1998}, which is a standard initialization method~\citep{babuschkinDeepMindJAXEcosystem2020}, and known to converge to the NTK limit in the infinite-width regime~\citep{liuLinearityLargeNonlinear2020}. 
For a more detailed description of the experimental setups, please refer to Appendix~\ref{sec:variance_scaling}.

% Describe the Goal of this section
The analysis and discussion of the results are carried out in Section~\ref{sec:scaling_mechanisms_scaling}.
The left column of Figure~\ref{fig:internal_scaling_mechanisms} presents the corresponding loss curves for scaling data and model size for the standard vision datasets.
To analyze the scaling behavior of NNs through the lens of the neural tangent kernel (NTK), we first aim to identify which aspects of the learning process are primarily affected by scaling the model and data size.
To do this, we analyze the dynamics of the NTK trace and the effective rank in~\cref{sec:dynamics_in_variance_limited_regime} for key quantities that characterize the learning process. 
We then compare the scaling behavior of key quantities in~\cref{sec:scaling_mechanisms_scaling} for model and data scaling.

\subsection{Dynamics and Key Quantities} \label{sec:dynamics_in_variance_limited_regime} %\looseness=-1
\begin{figure}[h]
    \begin{center}
    \includegraphics[width=\linewidth]{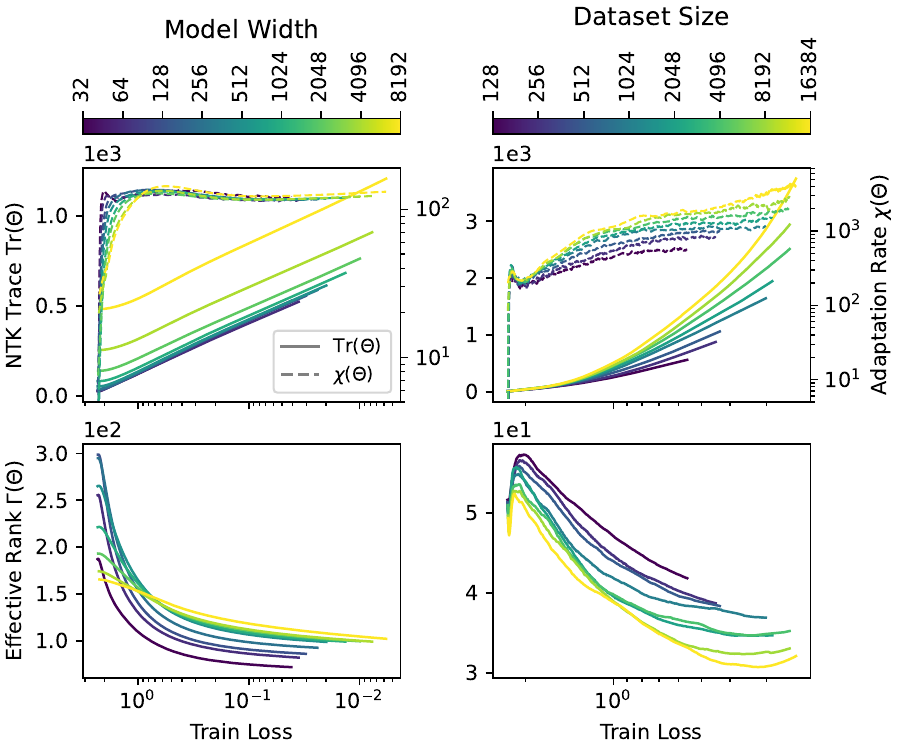}
    \end{center}
    \caption{
        Dynamics of the NTK trace $\text{Tr}(\bm{\Theta})$, adaptation rate $\chi (\bm{\Theta})$ and effective rank $\Gamma(\bm{\Theta})$
        as a function of training loss on the MNIST dataset.
        The figure shows learning curves for various model and dataset sizes, spanning multiple scales. 
        Each curve tracks the evolution from the initial model state to the point of minimum test loss (reading from left to right).
        Results are averaged over 20 ensembles.
    }
    \label{fig:dynamics_in_variance_limited_regime_mnist}
\end{figure}

% Describe the Goal of this section
To make the analysis of the NTK dynamics more accessible, we present the results for a representative example in Figure~\ref{fig:dynamics_in_variance_limited_regime_mnist}.
The results for the other datasets are presented in Appendix~\ref{sec:app_results_scaling_experiments} and lead to similar conclusions. 
% Trace Observations

\textbf{Trace Dynamics} \\
From Figure~\ref{fig:dynamics_in_variance_limited_regime_mnist}, we observe that the NTK trace increases continuously during training for both, scaling model and data sizes. 
% % Observations Model Scaling
% Rate of change
In the \textbf{model scaling} regime (top left panel of Figure~\ref{fig:dynamics_in_variance_limited_regime_mnist}),
we notice that after an initial phase, the trace increases linearly with the logarithm of the training loss.
% Interpretation Model Scaling
% From Section~\ref{sec:mechanistic_interpretation_ntk_quantities}, we can interpret the trace of the NTK as a measure of response to the training process.
% The observed behavior can be understood as follows: 
This means that irrespective of how far a model is trained, a (logarithmic) improvement in training loss is accompanied by a constant increase in trace. 
To quantify the rate at which this occurs, we introduce the trace adaptation rate,
\begin{equation}
    \chi(\bm{\Theta}) = \frac{\delta \text{Tr} \bm{\Theta}}{\delta \log \mathcal{L}_\text{train}} ,
\end{equation}
where $\mathcal{L}_\text{train}$ is the training loss.
We depict the resulting adaptation rate\footnote{
    The intuition behind the adaptation rate is to visualize the slope trace dynamics. Details on the computation of the adaptation rate can be found in Appendix~\ref{sec:adaptation_rate_computation}.
} together with the trace dynamics, in Figure~\ref{fig:dynamics_in_variance_limited_regime_mnist}.
After an initial phase, the adaptation rate appears to be invariant to scale and dynamics, and the differences induced by scaling model size are only attributed to the initial trace.
% This is in contrast to the initial trace, which scales with model size.\\
In the \textbf{data scaling} regime (top right panel of Figure~\ref{fig:dynamics_in_variance_limited_regime_mnist}), 
we observe that all models start with a similar trace, and as training progresses, the trace grows faster relative to the training loss. 
The increasing adaptation rate $\chi(\bm{\Theta})$ represents the same effect:
% This means that the response a model can build up depends on the training it has undergone so far.
% This indicates that a model's ability to respond and adapt is influenced by the training it has already undergone.
The further a model is in training, the larger its adaptation rate $\chi(\bm{\Theta})$. 
In contrast to scaling the model size, the adaptation rate depends on the amount of training the model has undergone. 
This allows us to use it as a proxy for quantifying how much the trace is increasing throughout training. \\
From our empirical observations, we conclude that there are two key properties that quantify the dynamics:
the trace at the initial state and the adaptation rate at the point of minimum test loss.
We will utilize these properties in Section~\ref{sec:scaling_mechanisms_scaling} to compare data and model scaling behavior.

\textbf{Effective Rank Dynamics} \\
% Observation from Dynamics
In contrast to the NTK trace dynamics, the effective rank (bottom panels of Figure~\ref{fig:dynamics_in_variance_limited_regime_mnist}) appears to decrease during large parts of training.
Over the course of training, we observe a clear ordering in the effective rank dynamics:
In the model scaling regime, training yields larger effective ranks for larger models, while in the data scaling regime, the effective rank decreases with increasing dataset size.
% Scaling model size
In order to capture the dynamics-induced ordering, we will utilize the effective rank at the point of minimum test loss $\Gamma(\bm{\Theta}_\text{min})$ as our reference point. 
% It quantifies the number of dominant eigenmodes the model resolves on the test data when performing best.

\subsection{Scaling} \label{sec:scaling_mechanisms_scaling} %\looseness=-1
\begin{figure*}[h]
    \begin{center}
    \includegraphics[width=\linewidth]{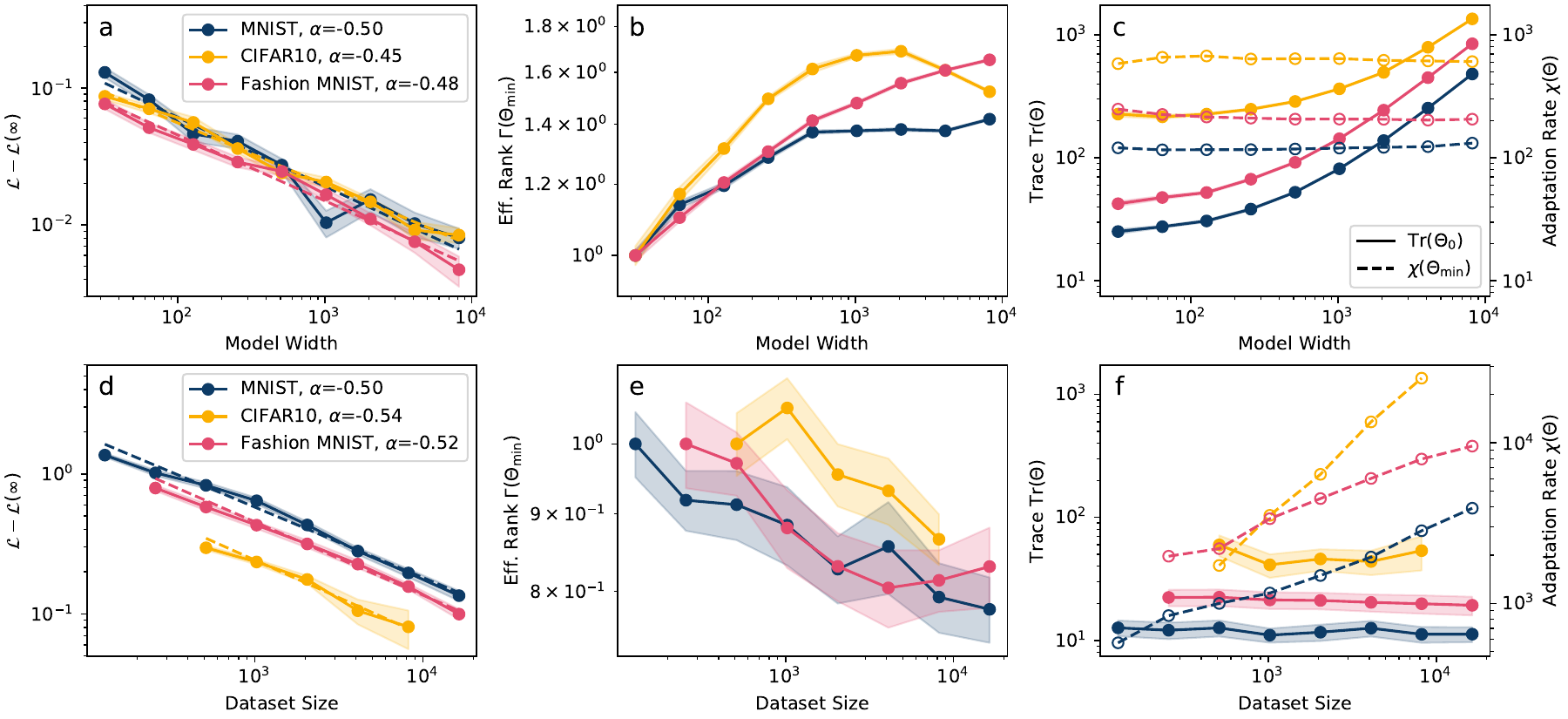}
    \end{center}
    \caption{
        Comparison of model scaling (a–c) and data scaling (d–f) for standard vision datasets. 
        In addition to test loss scaling, three key quantities characterizing the intrinsic network dynamics are shown: 
        the effective rank at minimum test loss $\Gamma (\bm{\Theta}_\text{min})$, the initial NTK trace $\text{Tr} (\bm{\Theta}_0)$, and the adaptation rate at minimum test loss $\chi (\bm{\Theta}_\text{min})$. 
        To improve visualization, the effective rank is normalized by its value for the smallest model or dataset size, respectively.
        Results are averaged over 20 ensembles, with shaded regions indicating the standard error.
    }
    \label{fig:internal_scaling_mechanisms}
\end{figure*}

% Introducing the Scaling Regimes
Figure~\ref{fig:internal_scaling_mechanisms} compares the scaling behavior of data scaling and model scaling for standard vision datasets.
In addition to the loss scaling, we present the key dynamic quantities derived in the previous section:
The NTK trace at initialization $\text{Tr}(\bm{\Theta}_0)$, the adaptation rate at the point of minimum test loss $\chi(\bm{\Theta}_\text{min})$, and the effective rank at the point of minimum test loss $\Gamma(\bm{\Theta}_\text{min})$.

%%%% Loss Scaling
% General Observations
From the loss scaling behavior, we observe that the performance of the models scales with a power law in both model and data sizes. 
Both scaling exponents are approximately equal to 0.5 and are consistent across different datasets and scaling regimes.
This is in line\footnote{
    We observe similar scaling exponents for model and data size, however, we were not able to reproduce the scaling exponent of 1.0 as reported in~\citet{bahriExplainingNeuralScaling2024b}.
    This discrepancy might be due to different learning rates and initializations utilized in the experiments.
} with the findings of~\citet{bahriExplainingNeuralScaling2024b}.

%%%% Effective Rank Scaling
% Observations Expansion
From Figure~\ref{fig:internal_scaling_mechanisms}, we observe that scaling the \textbf{model width} leads to an overall increase in the effective rank\footnote{
    The convergence and slight decreases in the effective rank for larger models
    will be discussed in Section~\ref{sec:feature_kernel_transition}.
    There, we find this to be a result of the model approaching the NTK limit.
}.
% Interpretation Expansion
% Following the insights from Section~\ref{sec:effective_rank_feature_learning},
% this suggests that when scaling the width \textit{additional capacity is utilized to refine more features}. 
% An increasing effective rank if the NTK indicates that more dimensions are effectively used when the model learns about the test data. 
This means that increasing a network's width, its extra capacity is used to include more dimensions into the learning process. 
% Data Scaling 
% Observations
By contrast, in the \textbf{data scaling} regime, we observe the opposite trend: the effective rank decreases as the dataset grows.
% Interpretation
% This can be understood as \textit{more data is utilized to compress internal features}.
% Intuition
% Intuitively, one can argue that since the model for data scaling is chosen very small, its capacity is smaller than the data complexity, which prevents the model from resolving the task perfectly.
The latter appears to be rather counterintuitive, as feeding a model more data might supply more information, i.e., directions for learning, which in turn should increase the effective rank. 
Yet we observe the opposite.
This apparent contradiction is best explained by the model's limited capacity:
Our data-scaling experiment uses a deliberately small network, so the dataset contains more information than the model can represent. 
In such an under-parameterised setting, the optimal model must compress the task to its most essential aspects.
The observed decrease in effective rank therefore, indicates that, for small models, discarding less-useful directions is more beneficial than trying to capture them all.
% This could the reason why the effective rank decreases:
% Increasing the training data increases the information density the model can access to learn the task.
% This could allow the model to better identify the relevant directions to learn. 

%%%%% Trace and Adaptation Rate 
The right panels of Figure~\ref{fig:internal_scaling_mechanisms} depict the trace at initialization and the adaptation rate at the point of minimum test loss.
% Model Scaling
    % Observations
In the \textbf{model scaling} regime, we observe that the initial trace increases with model width, while the adaptation rate remains constant.
    % Interpretation
This suggests that additional model capacity is used to increase the initial trace while the trace dynamics remain invariant to scale.
% This suggests that when scaling the model width, the initial response is increased, but the rate at which the network adapts to the task does not scale.
% Data Scaling
    % Observations
By contrast, in the \textbf{data scaling} regime, we observe the opposite behavior: the initial trace remains constant, while the adaptation rate increases with data size.
    % Interpretation
This indicates that more data enhances the dynamics and wider models the initial state of the NTK trace.

% Combining with performance
Combining this with our insights from the scaling experiments, we conclude the following, \textbf{opposing effects}: 
\begin{itemize}
    \item Scaling the width for large models, performance is enhanced with
    % \textbf{increasing initial network response and refinement of features}.
    \textbf{increasing the effective dimension of learning and scaling the initial NTK trace}.
    \item Scaling data for small models, performance is enhanced with
    % \textbf{increasing rate of response adaptation and compression of features}.
    \textbf{reducing the effective dimension of learning and scaling the NTK trace dynamics}.
\end{itemize}
\subsubsection{Discussion} %\looseness=-1
% Why should anyone care?
The scaling regime investigated here chooses either the data or the model comparably small while scaling the other factor large.
\citet{bahriExplainingNeuralScaling2024b} describes the scaling of these regimes to be of similar nature:
The bottlenecking factor controls scaling in the same way, regardless of whether the scaled property is data size or model size. 
% This dominant bottleneck contributes to performance scaling in a consistent way.
Our findings, however, reveal that while the performance scaling may appear similar, 
the intrinsic effects driving the model's behavior can differ fundamentally.  
This does not nessecarily contradict \citet{bahriExplainingNeuralScaling2024b}'s results, as their work focuses on the quantitative limitations of scaling laws. 
Instead, our work complements theirs by 
demonstrating that even when a dominant bottleneck comparably limits performance, the underlying model dynamics may still vary significantly.
% This highlights the importance of understanding not only the size of scaling contributions but also the mechanisms behind them. 
Working towards a comprehensive understanding of neural networks, it seems to be crucial to go beyond investigating the scaling of a single quantity. This work presents a step in this direction. \\
% Such insights into neural scaling are essential for connecting scaling theory to heuristic principles that aid applications. 
% The nature of these mechanisms can provide guidance for optimizing models more effectively, reducing model sizes, and better understanding the requirements for generalization.\\
The relations of the here presented findings to alignment-based theories are discussed in Appendix~\ref{sec:alignment_theories}. 
We find that our results appear to be independent of alignment mechanisms in neural networks.

\section{Feature-Kernel Transition} \label{sec:feature_kernel_transition} %\looseness=-1
The behaviour of infinitely wide neural networks is well characterised~\citep{jacotNeuralTangentKernel2018, yangFineGrainedSpectralPerspective2020, yangTensorProgramsII2020, yangFeatureLearningInfiniteWidth2022}.
In the kernel (or “lazy”) limit, training linearises around the random initialization, and the entire learning dynamics are dictated by a single static object—the neural tangent kernel (NTK).
Consequently, optimization converges to kernel regression: The network no longer learns representations but instead fits the data by re-weighting a fixed set of NTK basis functions~\citep{yangFeatureLearningInfiniteWidth2022}.
All effective features are thus predetermined at initialization, rather than being shaped by the training data, as for finite-width networks. \\
This transition from feature learning to kernel regression might be a key aspect to understanding the scaling behavior of neural networks.
It currently remains unclear at what width this occurs, and how training behavior changes.
The transition is of particular interest for understanding how large LLMs can possibly become before they lose their feature learning capabilities. 
To address this, we will investigate the effect of width scaling in a low-data regime, where it is computationally feasible to empirically study the transition. 
Details to the experimental setup are provided in Appendix~\ref{sec:variance_scaling}.
Besides allowing for studying fundamental aspects of scaling, this setting is particularly relevant for developing scientific foundation models, where data accessibility is often limited~\citep{smithAstronomiaExMachina2023,batatiaFoundationModelAtomistic2024,parkerAstroCLIPCrossModalFoundation2024}.

We enable studying this transition by extending the NTK-based quantities (NTK trace and effective rank) introduced in Section~\ref{sec:ntk_decomposition}:
Motivated by~\citet{chizatLazyTrainingDifferentiable2019b}, we will analyze the change of the NTK to quantify the transition. 
We use the trace as an intuitive way to approach this:
It is known that in the infinite-width limit, the NTK becomes static~\citep{jacotNeuralTangentKernel2018}, which means that the trace does not change during training. 
Consequently, we expect $\text{Tr}\bm{\Theta}_{\text{min}} - \text{Tr}\bm{\Theta}_{0}$ to converge to zero, where $\text{Tr}\bm{\Theta}_{\text{min}}$ is the trace at minimum test loss and $\text{Tr}\bm{\Theta}_{0}$ is the trace at initialization.
% This directly connects to the analysis performed in Section~\ref{sec:scaling_mechanisms_scaling}, where we used the adaptation rate $\chi(\bm{\Theta})$ (at minimum test loss) to quantify the change in NTK trace during training.
To allow for a comparison across different model sizes (and architectures), we remove the scale by defining the dimensionless \textit{trace ratio}:
\begin{align}
    \beta &= \left( \text{Tr}\bm{\Theta}_{\text{min}} - \text{Tr}\bm{\Theta}_{0} \right)/ \text{Tr}\bm{\Theta}_{0} 
    \label{eq:trace_ratio}
\end{align}
It measures the relative contribution of the change in NTK trace $\text{Tr}\bm{\Theta}_{\text{min}} - \text{Tr}\bm{\Theta}_{0}$ gained through training, normalized by the initial value $\text{Tr}\bm{\Theta}_{0}$. 
% The enumerator can equivalently be calculated by integrating the adaptation rate $\chi(\bm{\Theta})$ over the training loss dynamics $\left( \text{Tr}\bm{\Theta}_{\text{min}} - \text{Tr}\bm{\Theta}_{0} \right) = \int \chi(\bm{\Theta}) d (\log \mathcal{L}_\text{train})$.
% In contrast to the qualitative analysis from Section~\ref{sec:scaling_mechanisms_scaling}, we will here focus on the quantitative behavior by taking into account the full dynamics. 
We expect the limiting ratio to be $\lim_{\text{width} \to \infty} \beta = 0$.
\begin{figure*}[h]
    \begin{center}
    \includegraphics[width=1\linewidth]{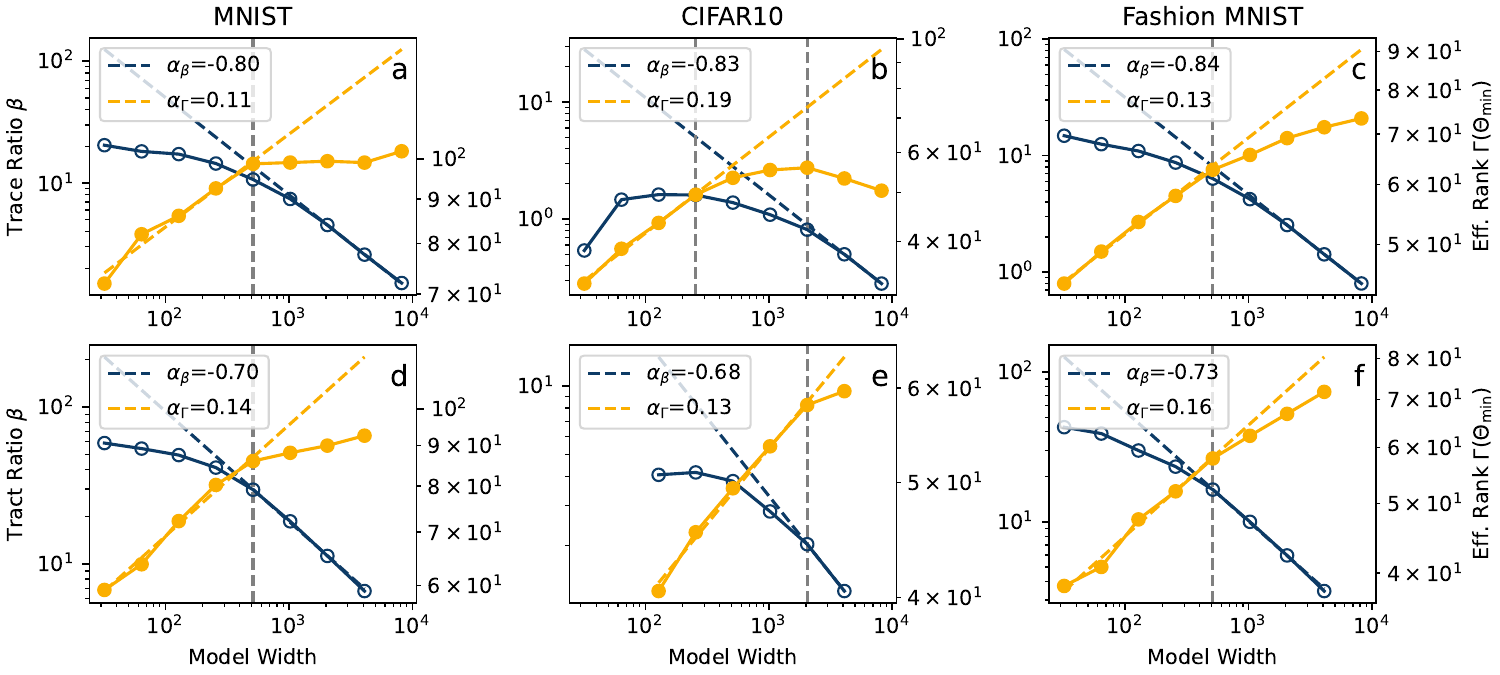}
    \end{center}
    \caption{
        Model size scaling of NTK trace ratio $\beta = \left( \text{Tr}\bm{\Theta}_{\text{min}} - \text{Tr}\bm{\Theta}_{0} \right)/ \text{Tr}\bm{\Theta}_{0} $  (empty markers) and effective rank at minimum test loss $\Gamma(\bm{\Theta}_\text{min})$ (filled markers), where $\text{Tr}\bm{\Theta}_{\text{min}}$ is the trace at minimum test loss and is the trace at initialization $\text{Tr}\bm{\Theta}_{\text{0}}$.
        The upper row (a-c) shows results for a three-layer dense neural network, while the lower row (d-f) corresponds to a four-layer dense neural network.
        The dashed lines represent fitted trends, and the vertical gray line indicates the transition from feature-driven to kernel-driven scaling. 
        The latter marks the maximum model width beyond which kernel-like behavior disrupts feature learning. 
        The exponents $\alpha_\beta, \alpha_\Gamma$ represent the scaling behavior of the trace ratio and the effective rank, respectively. 
        Results are averaged over 20 ensembles.
    }
    \label{fig:scaling_phases}
\end{figure*}
In addition, we will study the effective rank at minimum test loss $\Gamma(\bm{\Theta}_\text{min})$ as the second key quantity.
Because the effective rank summarises the eigenvalue spectrum of the NTK, it serves as a scale-invariant metric that is directly comparable across networks of different sizes.
In the infinite-width limit, the NTK becomes static, and we expect the effective rank to converge to a constant value $\lim_{\text{width} \to \infty} \Gamma(\bm{\Theta}_\text{min}) = \Gamma_\infty$.
Figure~\ref{fig:scaling_phases} presents the behavior of the trace ratio $\beta$ and the effective rank $\Gamma(\bm{\Theta}_\text{min})$ for width scaling of two distinct model depths.

\paragraph{Declining Trace Ratio as a Sign of Kernel Behavior \\} %\looseness=-1
% Model Scaling Observation
In all examples, we observe that $\beta$ decreases when the model size is increased, which corresponds to our expectation of converging to the kernel limit. 
% From the construbtion of $\beta$, this suggests that for wider models, the initial trace becomes the dominant factor.
By construction, the trace ratio measures the relative growth of the NTK trace during training; therefore its decline indicates that, for wider models, the initial trace increasingly dominates while the trace accrued through learning becomes negligible.
% For each model architecture, this decrease seems to converge to a power law, with nearly 
% similar exponents across the datasets.
% % Data Scaling Interpretation
% This implies the following:
% As we scale the width, the relevance of dynamics starts to diminish 
% at the same rate. 
% We 
We further identify that the decline takes place at the same rate:
${\beta \propto m^{\alpha_\beta}}$ for model width $m$ and scaling exponent ${\alpha_\beta \approx -0.8}$ for the shallower and ${\alpha_\beta \approx -0.7}$ for the deeper architecture.
% Impact
These observations suggest that, much like the limiting kernel function, the decline in NTK trace may be a common property of a given architecture, largely unaffected by the specific task under consideration. 

\paragraph{Effective Rank Scaling as a Sign of Feature Learning \\} %\looseness=-1
Analyzing the effective rank scaling, from all plots we observe that for narrow models it increases with a power law $\Gamma(\bm{\Theta}_\text{min}) \propto m^{\alpha_\Gamma}$ in model width $m$ and a scaling exponent $\alpha_\Gamma$. 
% Interpretation Increase
An increasing effective rank indicates that the additional capacity provided by wider models is utilized to effectively include more dimensions into the learning process.
The fact that this increase follows a power law suggests that if we double the width, we increase the number of effective dimensions by the same amount. 
% Taking into account the explanation provided in Section~\ref{sec:effective_rank_feature_learning},
% we understand additional capacity in this regime to be utilized to refine internal model features. 
% Observation Saturation and Interpretation
Beyond a certain width, we further observe that this trend plateaus—and may even reverse slightly—showing that once the network is sufficiently wide, adding more parameters yields diminishing returns in terms of new usable directions. \\
We expect smaller models to adapt to the data by learning feature representations~\citep{yangFeatureLearningInfiniteWidth2022}.
Since the effective rank scaling consistently occurs throughout the investigated datasets, we interpret the power-law increase to be a sign of feature learning. % and the plateau to be a sign of saturation.
We include a more detailed analysis of how the effective rank captures feature learning in Appendix~\ref{sec:effective_rank_feature_learning_app}.

\paragraph{Identifying the Maximum Model Width for Feature Learning \\} %\looseness=-1
% Compare to Response Ratio
Figure~\ref{fig:scaling_phases} further compares the two NTK quantities: 
The saturation of the effective rank to the decline of the trace ratio.
In most examples\footnote{
    The obvious deviating behavior in Figure~\ref{fig:scaling_phases} b, is discussed in Section~\ref{sec:discussion_fkt}. 
}, we can observe that the effective rank starts saturating at the same point where the trace ratio $\beta$ starts to decrease.
% Conclusion
% This suggests that the ability of feature learning starts to decline when the kernel regime becomes dominant.
We obtain a transition in model width separating two regimes:
Below, the effective rank increases predictably with width, suggesting active feature learning. 
Above this width, further scaling diminishes the relevance of trace dynamics, indicating a drift toward the kernel regime.
% Below this transition, the effective rank scales (predictably) with the model width, 
% while above the transition, scaling yields emphasises on the initial trace, indicating that the model is approaching the kernel limit.
We interpret these cross-overs to mark the transition from feature-driven scaling in small models to kernel-driven scaling in large models.
Therefore, we conclude that a network operating at the transition marks the maximum model width that supports feature-driven learning.

We identify this model width for networks trained on MNIST and Fashion-MNIST at a width of about 500 and for CIFAR-10 with a width of 1000. 
% As we train deep feed-forward neural networks, the identified sizes do not directly generalize to other architectures such as transformers used in LLMs. 
% However, since dense feed-forward layers are an essential part of the transformer architecture~\citep{vaswani17a-pre}, we can compare to the feed-forward layer widths in state-of-the-art LLMs which for example are 14336 in Mistral-7B~\citep{gupta25a-pre} or between 18432-73728 in the Gemma series models~\citep{team24a-pre}. 
% This is more than one order of magnitude larger than the identified model width for feature-driven scaling in our experiments.
% In that regard, our study shows the potential relevance of kernel-like behavior for LLMs, since we demonstrate that kernel-like behavior can occur in models significantly shallower than state-of-the-art LLMs.
% We further present the starting point to disentangle effects that potentially shift the transition point to larger model sizes, such as the influence of the attention layers and large training data sets.
As we train deep feed-forward neural networks, these sizes do not directly generalize to other architectures, such as transformers used in LLMs.
However, since dense feed-forward layers are an essential part of transformer architectures, we can compare the widths of the feed-forward layers in state-of-the-art LLMs. For example, the width is 14336 in Mistral-7B~\citep{gupta25a-pre}, and it ranges from 18432 to 73728 in the Gemma series models~\citep{team24a-pre}. 
These values are more than one order of magnitude larger than the widths we identified for feature-driven scaling in our experiments.
Our study shows the potential relevance of kernel-like behavior for LLMs since we demonstrate that this can occur in models that are significantly shallower than state-of-the-art LLMs.
Furthermore, we present a starting point for disentangling the effects that potentially shift the transition point to larger model sizes, such as the influence of other architectural components and large training datasets.

% We identify this model size for networks trained on MNIST and Fashion-MNIST at a width of about 500 and for CIFAR-10 with a width of 1000, which is far below what is utilized in LLMs~\citep{brownLanguageModelsAre2020}.
% In terms of parameters, this corresponds to models with approximately 0.7 million and 4.2 million parameters, respectively.
% This parameter scale is of similar magnitude to the smallest models used in language modeling tasks, 
% and several orders below the typical sizes of LLMs~\citep{cohnEELBERTTinyModels2023}. 
% This demonstrates that kernel-like behavior can occur in models significantly smaller than state-of-the-art LLMs, suggesting its potential relevance across a wider range of model scales and applications, and thus warranting further investigation.

% Although the dataset size used in this study is relatively small, our results show that the effects of kernel behavior can be identified in models with less than one million parameters. 

\subsection{Discussion} \label{sec:discussion_fkt} %\looseness=-1
% \paragraph{Transition in Effective Rank and Trace Ratio \\} %\looseness=-1
In Figure~\ref{fig:scaling_phases} b, the shallower architecture is trained on CIFAR-10.
It is clearly visible that the transition in effective rank is reached before the transition in trace ratio. 
Taking a closer look at the other datasets trained on the shallower model (Figure~\ref{fig:scaling_phases} a-c), we observe a similar trend, but less pronounced.
% The trace ratio $\beta$ quantifies change in scale of the NTK, while the effective rank $\Gamma(\bm{\Theta}_\text{min})$ measures an aspect of the kernel distribution.
Note that the trace ratio $\beta$ and the effective rank $\Gamma$ quantify distinct aspects of the learning dynamics.
It appears that these different aspects can be affected at different scales.
% While the trace ratio measures the dynamically accumulated change in the NTK, it provides us with a measure of "how much feature learning is still possible".
% The effective rank, on the other hand, quantifies the number of dimensions that are effectively used to resolve the task. 
% It could be understood as an "aspect of feature learning that actually takes place". 
% In this sense, it is not surprising that the effective rank can saturate before the trace ratio starts to decline. \\
For the deeper architecture (Figure~\ref{fig:scaling_phases} d-f), however, we do not observe this difference in the transition of effective rank and trace ratio:
Increasing model depth, both NTK quantities transition at the same width.
% indicating feature-driven scaling in the effective rank can be maintained until the trace ratio starts to decrease.
This suggests that the depth of the network plays a non-negligible role in the transition from feature-driven to kernel-driven scaling.

While in most of our experiments the transitions roughly coincide, the minimum model width for feature-driven scaling appears to demand for a more differentiated picture. 
Depth dependence is most pronounced on the most complex dataset, CIFAR-10.
Extrapolating from our results to state-of-the-art tasks, we would scale up data-complexity by increasing the dataset size and switching to a more complex dataset.
In this scenario, we expect the difference in transition points to become more pronounced.
% Even if using larger and more complex models (like ResNets and Vision Transformers) might counteract this effect, 
Understanding all facets of the transition seems to require investigating multiple transition points between feature and kernel learning.

\section{Conclusion}
\subsection{Summary}
This work provides a starting point for understanding scaling behavior in neural networks by empirically analyzing neural network dynamics.
By studying the trace and the effective rank of the neural tangent kernel, we analyzed scaling effects beyond the loss scaling law. 
We addressed two questions:
1. How does scaling model size and dataset size intrinsically affect a model’s behavior?
2. At what model size can we identify the influence of the kernel regime? \\
% Scaling Mechanisms
We addressed the first question by empirically comparing the scaling of model and data size.
We observed similar scaling exponents for performance (loss), but fundamentally different effects on the internal model dynamics:
Effective rank and NTK trace dynamics are affected in opposite ways.
The effective rank increases with model size, while it decreases with data size. 
Scaling the model size affects the initial NTK trace, while scaling the dataset size affects the trace dynamics. 
These results highlight that focusing solely on loss scaling exponents may overlook critical differences in the underlying principles. \\
% Feature Kernel Transition
To address the second question, we investigated the transition from feature-driven to kernel-driven scaling. 
This was achieved by analyzing the trace ratio (a measure of the amount of kernel dynamics, see Equation~\ref{eq:trace_ratio}) and effective rank as a function of model width.
By studying their scaling behavior, we identified transition points in both quantities, which coincide for sufficiently deep networks.
% We identified a transition point where the network shifts from feature-driven to kernel-like behavior.
% We identified a transition at finite model size where the network shifts from feature-driven to kernel-like behavior.
This transition arguably marks the maximum model beyond which feature learning begins to diminish. 
The identified model width are far below the typical widths of large language models, building a foundation for addressing the question of whether kernel-like behavior affects the performance of state-of-the-art models. 
A closer look at the transition revealed that the model depth affects the coincidence of the transition points in trace ratio and effective rank.
This suggests that the interplay of width and depth is crucial for a comprehensive understanding of the feature-kernel transition.

\subsection{Outlook} \label{sec:outlook}
% Outlook Mechanisms -> Qualitatively reconstruct scaling exponents
A complete understanding of neural scaling laws requires bridging theoretical models with empirical observations.
Such a connection can be achieved by identifying the underlying effects that influence the performance scaling of models, for which this work provides a starting point.
% % What to do in order to connect the mechanisms to the scaling exponents?
% Towards constructing such an effective theory of scaling, we suggest the following:
% 1. Construct a set of observables that can capture relevant dynamic effects by including directional NTK quantities, such as kernel alignment and kernel distance. 
% 2. Quantitatively reconstruct scaling exponents from these observables. 
% % Quantitative Reconstruction
% This can yield a way to quantify the relevance of different intrinsic model effects and their contributions to scaling behavior. \\
In addition to the here presented results, other mechanisms underlying NN training have been identified through measures such as kernel distance and kernel alignment~\citep{fortDeepLearningKernel2020, baratinImplicitRegularizationNeural2021, chizatLazyTrainingDifferentiable2019b}.
Identifying and combining all relevant observables into a set of measurements, one can aim to reconstruct scaling laws from that set. 
This could provide a way to develop an effective scaling theory. \\
% FKT Outlook
Our results on transitioning from feature to kernel learning highlight the potential relevance of kernel-like behavior for LLMs. 
From the discussion in Section~\ref{sec:discussion_fkt}, we suspect that different aspects of the learning dynamics may be affected at different scales. 
In order to understand all facets in which kernel-like behavior can influence performance in state-of-the-art models, it seems crucial to consider the effect of depth in future studies as an essential architectural hyperparameter.

\subsection{Limitations}
% Limitations wrt to the variance-limited regime
A limitation of the results presented here is that they focus on the low-data and small-model regimes. 
The dynamics shown in Figures~\ref{fig:dynamics_in_variance_limited_regime_mnist},~\ref{fig:network_response}, and~\ref{fig:effective_rank} may not directly be comparable to scenarios where the model and data have more similar levels of complexity. 
In such cases, the interaction between model capacity and data complexity could lead to different learning dynamics that are not captured in the current analysis.
% Limitation of the completeness of the observables. 
Moreover, breaking down dynamics into two observables (trace and effective rank of the NTK) may not be sufficient to capture the full complexity of the learning process. 
While this does not pose a significant issue for our analysis, as we did not aim to quantitatively reconstruct scaling exponents, it becomes a limitation if the latter is the ultimate goal. 
To achieve a more comprehensive understanding and to accurately capture all aspects of neural network dynamics, it might be necessary to extend the studied observables in future work, as discussed in Section~\ref{sec:outlook}.
% Limitations of simplified dynamics
Our analysis further relied on comparing NTK quantities at specific points in training (at initialization and at minimum test loss). 
It has to be tested whether this simplification holds across regimes closer to the interpolation threshold. 
% These simplifications have to be tested to hold across regimes closer to the interpolation threshold.
The analysis should then be adapted to take into account the potentially more complex dynamics in these regimes.

\section*{Software and Data}

All experiments were conducted with JAX~\citep{babuschkinDeepMindJAXEcosystem2020}. 
The NN models were implemented with Flax~\citep{heekFlaxNeuralNetwork2024},
and the NTK computations were carried out with Neural Tangents~\citep{novakNeuralTangentsFast2019b}.
% NTK trace and effective rank dynamics are computed utilizing ZnNL~\citep{ortiz-toveyPhenomenologicalUnderstandingNeural2023} and 
% papyrus~\citep{nikolaou_zincwarepapyrus_2025}.

% \section*{Impact Statement}
% This paper presents work whose goal is to advance the field of 
% Machine Learning. There are many potential societal consequences 
% of our work, none which we feel must be specifically highlighted here.

% % Acknowledgements should only appear in the accepted version.
\section*{Acknowledgements}

We thank the Deutsche Forschungsgemeinschaft (DFG, German Research Foundation) for supporting this work by funding - EXC2075 – 390740016 under Germany's Excellence Strategy. We acknowledge the support by the Stuttgart Center for Simulation Science (SimTech).
The authors thank the International Max Planck Research School for Intelligent Systems (IMPRS-IS) for supporting Konstantin Nikolaou. 
We acknowledge funding from the Deutsche Forschungsgemeinschaft (DFG, German Research Foundation) through the Compute Cluster grant no. 492175459.
% We want to thank


\bibliography{bibliography}
\bibliographystyle{icml2025}

%%%%%%%%%%%%%%%%%%%%%%%%%%%%%%%%%%%%%%%%%%%%%%%%%%%%%%%%%%%%%%%%%%%%%%%%%%%%%%%
%%%%%%%%%%%%%%%%%%%%%%%%%%%%%%%%%%%%%%%%%%%%%%%%%%%%%%%%%%%%%%%%%%%%%%%%%%%%%%%
% APPENDIX
%%%%%%%%%%%%%%%%%%%%%%%%%%%%%%%%%%%%%%%%%%%%%%%%%%%%%%%%%%%%%%%%%%%%%%%%%%%%%%%
%%%%%%%%%%%%%%%%%%%%%%%%%%%%%%%%%%%%%%%%%%%%%%%%%%%%%%%%%%%%%%%%%%%%%%%%%%%%%%%
\newpage
\appendix
\onecolumn

\section{NTK Decomposition} \label{sec:ntk_decomposition_app}

Let $f_k(\vx, \vtheta)$ be the $k-$th dimension of the prediction of an NN 
with parameters $\vtheta$ and input $\vx$.
The network is trained on a dataset 
$\train = \{(\vx_i, \vy_i)\}_{i=1}^d = (\bm{X}, \bm{Y})$ with $d$ samples 
using loss function $\mathcal{L}(\vf(\bm{X}, \vtheta), \bm{Y}) = \frac{1}{d} \sum_{i} \ell(\vf(\vx_i, \vtheta), \vy_i)$.
Updating the network with gradient flow and continuous time $t$ the predictions evolve as
\begin{align}
    \dot{f}_k &= \frac{d f_k}{dt} 
    = - \nabla_{\vtheta} f_k(\bm{X}, \vtheta) \frac{d \vtheta}{dt}
    = - \bm{\Theta}_{kl} \, \nabla_{f_l} \mathcal{L}(f_l(\bm{X}, \vtheta), \vy) \\
    \bm{\Theta}_{kl} &= \nabla_{\vtheta} f_k(\bm{X}, \vtheta)^\top \nabla_{\vtheta} f_l(\bm{X}, \vtheta)
\end{align}
with $\bm{\Theta}_{kl}$ being the Neural Tangent Kernel (NTK) of the 
$k-$th and $l-$th output dimension. 
The full NTK is a rank 4 tensor $\bm{\Theta} \in \mathbb{R}^{d \times d \times n \times n}$, 
where $n$ is the number of output dimensions of the model. 

To decompose the NTK into its eigenvalues and eigenvectors, we reshape 
the NTK into a matrix $\bm{\Theta} \in \mathbb{R}^{d \cdot n \times d \cdot n}$,
and proceed as described in Section~\ref{sec:ntk_decomposition}.

\section{Effective Rank} \label{sec:effective_rank_app}

\subsection{Effective Rank and Entropy} \label{sec:effective_rank_entropy_app}

The effective rank introduced in Section~\ref{sec:ntk_decomposition} is defined 
in Equation~\ref{eq:effective_rank} via the von Neumann entropy of a matrix. 
As the von Neumann entropy is a measure of the correlation of a density matrix, the 
effective rank can be understood as a rescaling of the entropy to interpret it in the 
context of dimensionality measures. 
For all the scaling and dynamics analysis in this work,
the effective rank and the von Neumann entropy present exchangeable measures. 
We utilize the effective rank as it might provide the reader with 
a more intuitive interpretation than the entropy of a matrix.

\subsection{Effective Rank and Feature Learning} \label{sec:effective_rank_feature_learning_app}

%%%%%%%%%%%%%%%%%%%%%%%%%%%%%%%%%%%%%%%%%%%%%%%%%%%%%%%%%%%%%%%%%%%%%%%%%%%%%%%%%%%%%%%%%%%%%%%

% \section{Mechanistic Interpretation of NTK Quantities} %\looseness=-1 
\label{sec:mechanistic_interpretation_ntk_quantities}
% \subsection{NTK Trace and Network Response} %\looseness=-1
% \label{sec:mechanistic_interpretation_ntk_trace}
% The NTK represents the linear response of a neural network during training. 
% By examining the trace of the NTK, we capture the global magnitude of this response. 
% This offers an intuitive interpretation: the trace quantifies how responsive the network is to the training process. 
% Since the NTK is computed from the test data in our studies, the trace specifically reflects the extent to which the network responds to the underlying task.
% % Since the trace of the NTK is directly proportional to the change of the network function 
% % $\text{Tr} \bm{\Lambda} = \text{Tr} \bm{\Theta} 
% % % =  \| \nabla_{\vtheta} f(\bm{X}, \vtheta) \|^2
% % \propto \dot{f}$, 
% % it scales the rate at which a model trains. 
% % It hence measures the response contribution of the network to the training process.\\
% % which is why we refer to it as the \textit{network response}: 
% % \begin{equation}
% %     \Phi = \text{Tr} \bm{\Theta} \label{eq:speed}
% % \end{equation}
% \subsection{Effective Rank and Feature Learning} \label{sec:effective_rank_feature_learning} %\looseness=-1
% % Explain the Problem of Effective Rank
The effective rank of a matrix is a measure of how many effective dimensions a matrix 
consists of~\citep{royEffectiveRankMeasure2007}.
In the context of the NTK, it thus quantifies the number of dominant eigenmodes that 
contribute to the learning process.
% Explain the Solution
Investigating the effective rank over the course of learning and scaling, 
we can identify how the number of dominant eigenmodes changes. 
If it decreases, the model reduces the number of dimensions utilized to resolve the task, 
while an increase means that the network utilizes more dimensions. 
% In the first case, we can think of the network as \textit{compressing} learned structures
% while in the latter case, the network \textit{refines} learned structures to resolve the task. 

% Goal of this section
The change of dimensionality in learned structures can directly be related to the concept of feature learning.
% Theoretical Background
Below, in Section~\cref{sec:effective_rank_feature_learning_theory_app}, we provide the mathematical background for this interpretation.
From this, we identify the effective rank of the NTK \textit{to be directly related to the effective dimension in which feature learning takes place}. \\
% Experimental Setup
In order to numerically showcase this, we conduct the following experiment:
% \footnote{
%     A detailed description of the experiment can be found in \cref{sec:noise_scaling} 
%     and the corresponding performance plots are presented in \cref{fig:controlled_dynamics_acc}. 
% }:
% Training NNs on the CIFAR-4 dataset, we remove varying amounts of class-internal samples and replace them by noisy copies of samples that are left. 
Optimizing NNs on CIFAR-4\footnote{
    A version of CIFAR-10 consisting of the classes ”birds”, ”cats”, ”airplanes”, and ”automobiles”.
    More information on the dataset can be found in Section~\ref{sec:noise_scaling} and~\ref{sec:noise_scaling_results}.
}, for each class, we remove varying amounts of training samples and replace them with noisy copies of the samples that are left. 
This keeps the learning task the same but changes the amount of features a network can learn.
For each level of replacement, we examine the effective rank dynamics of the NTK during training.
% What do we expect
For the effective rank to capture the amount of learned features, we expect it to increase with the level of features present in the data. 

\begin{figure}[]
    \begin{center}
    \includegraphics[width=0.6\linewidth]{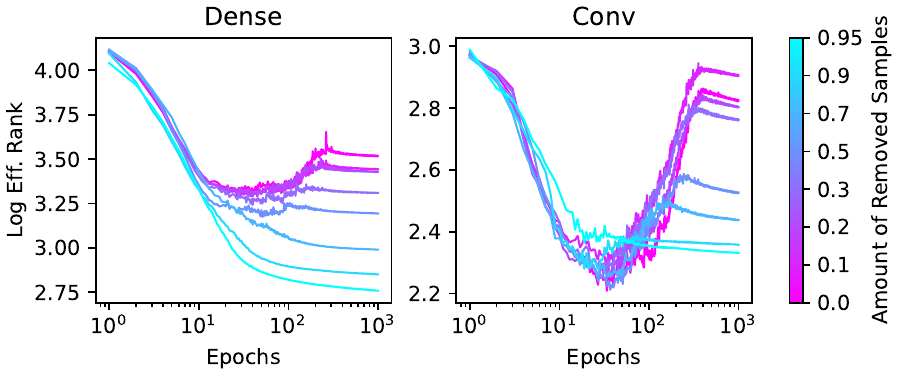}
    \end{center}
    \caption{
        Effective Rank $\Gamma(\bm{\Theta})$ dynamics of dense and convolutional NNs trained on CIFAR-4.
        We create datasets containing different amounts of features by removing varying numbers of training samples and replacing them with noisy copies of the remaining samples.
        A detailed description of the experiment can be found in \cref{sec:noise_scaling} 
        and the corresponding performance plots are presented in \cref{fig:controlled_dynamics_acc}. 
    }
    \label{fig:controlling_dynamics}
\end{figure}
From the results in Figure~\ref{fig:controlling_dynamics}, we observe a decreasing effective rank for short training times, followed by a varying amount of increase for longer training times.
% Utilizing the interpretation from above, we identify two distinct stages\footnote{
%         % Note that for the setup utilized here, the network and data complexity approximately match.
%         % This means that for an appropriately sized network, we can see the training dynamics 
%         % being shaped by compressing and refining internal structures. 
%         Identifying two stages is in line with the observations presented in~\citet{tovey_collective_2024}.
%         The behavior also aligns with the theory of NNs learning 
%         distributions of increasing complexity~\citet{refinettiNeuralNetworksTrained2023},
%         as we observe that information is first compressed and then refined. 
%         Starting from compressed internal representations, refinement could allow resolving for
%         information of increasing complexity. 
%     } in the training process:
This two-stage behavior is in line with the observations presented in~\citet{toveyCollectiveVariablesNeural2024} and aligns with the theory of NNs learning distributions of increasing complexity~\citet{refinettiNeuralNetworksTrained2023}.
% A stage of compressing learned structures, followed by a stage of refining learned structures.
We identify that the level of increase is controlled by the amount of class-internal features the model can learn, such that the more features are present, the further the effective rank can increase.
% We observe that the more class-internal features are present, the more the network refines its learned structures.
This empirically demonstrates that the effective rank is directly related to the amount of features learned by the network.
% that the amount of features a network learns can be captured by the effective rank.

\subsubsection{Theoretical Background} \label{sec:effective_rank_feature_learning_theory_app}

From the definition of the NTK in Equation~\ref{eq:ntk}, it can be understood as a data-data covariance matrix that describes the first order correlation of data in the network. 
At the same time, it similarly describes how the network learns from the data.
The effective rank of the NTK thus captures the effective dimensionality of learning and data correlation in the network.

In the context of feature learning, \textit{features} are usually understood as structured representations of the data that are learned by the network~\citep{lecun15a,yangFeatureLearningInfiniteWidth2022,bordelonSelfConsistentDynamicalField2022}.
We will use the terms features, structured representations, and activations interchangeably.
In the following, we will relate the feature-feature covariance matrix, the Conjugate Kernel (CK), to the NTK, establishing a link between NTK and feature learning. 
To do this, we will utilize a derivation from~\citet{fanSpectraConjugateKernel2020a} (Equations~\ref{eq:mlp_full} to~\ref{eq:structure_matrix}): \\
Consider a mutli-layer perceptron of the following form
% We consider a fully-connected, feedforward neural network with input dimension $d_0$, 
% hidden layers of dimensions $d_1, \ldots, d_L$, and a scalar output. For an input $\mathbf{x} \in \mathbb{R}^{d_0}$, we parametrize the network as
\begin{equation}
    f(\bm{x}, \bm{\theta})=\bm{w}^{\top} \frac{1}{\sqrt{w_L}} \sigma\left(\bm{W}_L \frac{1}{\sqrt{w_{L-1}}} \sigma\left(\ldots \frac{1}{\sqrt{w_2}} \sigma\left(\bm{W}_2 \frac{1}{\sqrt{w_1}} \sigma\left(\bm{W}_1 \bm{x}\right)\right)\right)\right) ,
    \label{eq:mlp_full}
\end{equation}
with activation function $\sigma$, parameters $\bm{\theta}=\left(\operatorname{vec}\left(\bm{W}_1\right), \ldots, \operatorname{vec}\left(\bm{W}_L\right), \bm{w}\right)$ and $w_\ell$ the width of the $\ell-$th layer.
Given $d$ training samples $\mathbf{x}_1, \ldots, \mathbf{x}_d $, we can define the matrices of activations $X_{\ell}$ at layer $\ell$ as
\begin{equation}
    \bm{X} \equiv \bm{X}_0=\left(\begin{array}{lll}
    \bm{x}_1, & \ldots &, \bm{x}_n
    \end{array}\right), \quad \bm{X}_{\ell}=\frac{1}{\sqrt{w_{\ell}}} \sigma\left(\bm{W}_{\ell} \bm{X}_{\ell-1}\right) 
\end{equation}
Utilizing this parametrization, we can define the Conjugate Kernel (CK) $\bm{C}_{\ell}$ of layer ${\ell}$
 as the gram matrix of the activations:
\begin{equation}
    \bm{C}_{\ell} = \bm{X}_{\ell}^\top \bm{X}_{\ell}
    \label{eq:ck}
\end{equation}
Likewise, and as introduced in Section~\ref{sec:ntk_decomposition},
the Neural Tangent Kernel (NTK) $\bm{\Theta}$ is defined as the gram matrix of 
gradients of the network predictions with respect to the weights:
\begin{equation}
    \bm{\Theta} = \nabla_{\bm{\theta}} f(\bm{X}, \bm{\theta})^\top \nabla_{\bm{\theta}} f(\bm{X}, \bm{\theta})
    \label{eq:ntk}
\end{equation}

Utilizing this definition, we can split each layer's contribution to the NTK into its 
conjugate kernel and an operator $\bm{S}_{\ell}$ which we will refer to as the sensitivity matrix:
\begin{equation}
    \bm{\Theta} = \bm{X}_L^\top \bm{X}_L + \sum_{\ell=1}^L \left( \bm{S}_{\ell}^\top \bm{S}_{\ell} \right) \odot \left( \bm{X}_{\ell-1}^\top \bm{X}_{\ell-1} \right)
    \label{eq:ntk_structure_decomposition}
\end{equation}
The sensitivity matrix $\bm{S}_{\ell}$ can be understood as how gradients propagate backwards through the network to the $\ell-$th layer.
Or, from the perspective of the CK, $\bm{S}_{\ell}^\top \bm{S}_{\ell} := \bm{\Sigma}_{\ell}$ describes how sensitive the output is to the $\ell-1-$th layer CK, which is why we refer to $\bm{\Sigma}_{\ell}$ as kernel sensitivity.
It conencts the relations of features in a layer to the learning dynamics of the network.
% how the $\ell-$th layer CK propagates through the network influencing changes in the output.
The $\alpha^{\text {th }}$ column of $\bm{S}_{\ell}$ is given by
\begin{equation}
    \bm{s}_\alpha^{\ell} = \bm{D}_\alpha^{\ell} \frac{\bm{W}_{\ell+1}^\top}{\sqrt{w_{\ell}}} \bm{D}_\alpha^{\ell+1} \frac{\bm{W}_{\ell+2}^\top}{\sqrt{w_{\ell+1}}} \ldots \frac{\bm{W}_L^\top}{\sqrt{w_{L-1}}} \bm{D}_\alpha^L \frac{\bm{w}}{\sqrt{w_L}} , 
    \label{eq:structure_matrix}
\end{equation}
with diagonal matrices $\bm{D}_\alpha^k = \operatorname{diag}\left(\sigma'\left(\bm{W}_k \bm{x}_\alpha^{k-1}\right)\right)$
and $\bm{x}_\alpha^{\ell}$ the $\alpha^{\text {th }}$ column of $\bm{X}_{\ell}$.
For more details on the derivation of this decomposition, we refer to~\citet{fanSpectraConjugateKernel2020a}. 

From Equation~\ref{eq:ntk_structure_decomposition} we can see that the NTK can be rewritten into a sum of the Conjugate Kernels at each layer. 
This formalizes a connection between the NTK governing the learning dynamics and the Conjugate Kernel (CK), which describes the feature-feature covariance matrix. 
This means that the relations (inner products) between features directly influence the learning dynamics of the neural network.
% For example, if the NTK increases its rank over training, there can be two principal mechanisms: 1) For similar kernel sensitivity, some layer's CK increases its rank. This corresponds to learning more orthogonal features. 2) For similar CKs, the kernel sensitivity shifts its focus from layers of low-rank to layers of high-rank CKs. This corresponds to selecting embeddings of more rich features to learn from.
Altering the structure of embeddings, by e.g., increasing the rank of layer activations, directly informs the NTK. 
The degree to which the NTK rank is affected by this is determined by the kernel sensitivity $\bm{\Sigma}_{\ell}$.
We conclude that the effective rank of the NTK is directly related to the effective dimension in which feature learning takes place.

\subsection{Effective Rank Computation}
Since we argue that the effective rank of the NTK is a measure of how many effective dimensions
learning takes place, let's consider the following example of a toy NTK
\begin{equation}
    \bm{\Theta}_1 = \begin{pmatrix}
        1/2 & 0 \\
        0 & 1/2
    \end{pmatrix}
\end{equation}
of two data points and a network of a single output dimension.
This example NTK shows the case in which training on the first data point does not influence the representation of the second data point; thus, training takes place in two effective dimensions.
This is reflected in the effective rank of the NTK, which is $\Gamma(\bm{\Theta}_1) = 2$. 
Considering a second example NTK 
\begin{equation}
    \bm{\Theta}_2 = \begin{pmatrix}
        1/3 & 0 \\
        0 & 2/3
    \end{pmatrix}
\end{equation}
of non-similar gradient magnitudes for the two data points, the effective rank is
$\Gamma(\bm{\Theta}_2) < 2$, as the two data points affect the network on different magnitudes. 
The underlying learning dynamics, however, take place on two independent dimensions, with
each eigenvalue scaling the rate of learning.
This is why, for the analysis of the effective rank, we consider the NTK of normalized 
gradient magnitudes to correct for the magnitude scaling of the gradients.

\subsection{Adaptation Rate Computation} \label{sec:adaptation_rate_computation}

Due to fluctuations in the trace dynamics, we calculate a running average of the mean trace and mean log training loss over 20 epochs. 
To derive the adaptation rate, we compute the numerical derivative of the time-averaged trace with respect to the time-averaged log train loss. 
Due to this computation, we do not provide error estimations for the adaptation rate \cref{fig:internal_scaling_mechanisms}.
% the adaptation rate is derived from the running average of the mean trace by computing the numerical derivative. 
% The trace is averaged over 20 ensembles and the running average is computed over 20 epochs.

\newpage
\section{Methods} \label{sec:methods}

\subsection{Evaluation Methods} \label{sec:evaluation_methods}
In this section, we comment on the evaluation methods utilized in this work, since we rely on the train and 
test loss as the main comparison measure in the scaling experiments presented 
most figures of this work.
% in Figures~\ref{fig:network_response},~\ref{fig:effective_rank},~\ref{fig:scaling_phases},~\ref{fig:loss_scaling},~\ref{fig:network_response_fc3},~\ref{fig:effective_rank_fc3}, and ~\ref{fig:alignment}.
% Explain the Problem of Network Response
Comparing dynamic quantities of NNs across different setups is challenging.
Applying a gradient step with learning rate $\eta$ in a billion-dimensional 
parameter space may behave very differently from a step of similar size in a million-dimensional space. 
Training runs with different learning rates, batch sizes, and architectures 
can, therefore, lead to vastly different training dynamics, making it impossible to compare
them at the same step or epoch.
% Explain the Solution
To avoid questionable comparisons,
% instead of relying on steps and epochs, 
we refer to the loss as the objective measure of inherent progress whenever epochs cannot be compared.
We can compare any observable of interest at the same train or test loss value, 
giving us measures for the progress of training or generalization. 

% \subsection{Computational Methods}
% All experiments are conducted with JAX~\citep{babuschkinDeepMindJAXEcosystem2020}. 
% The NN models are implemented with Flax~\citep{heekFlaxNeuralNetwork2024},
% and the NTK computations are carried out with Neural Tangents~\citep{novakNeuralTangentsFast2019b}.
% NTK trace and effective rank dynamics are computed utilizing ZnNL~\citep{ortiz-toveyPhenomenologicalUnderstandingNeural2023} and 
% % papyrus~\citep{nikolaou_papyrus_nodate}.
% XXX. 

\section{Experimental Setups} \label{sec:experimental_setups}

\subsection{Scaling Experiment} \label{sec:variance_scaling}
For the scaling results presented in 
% Figures~\ref{fig:network_response},~\ref{fig:effective_rank},~\ref{fig:scaling_phases},~\ref{fig:loss_scaling},~\ref{fig:network_response_fc3},~\ref{fig:effective_rank_fc3}, and ~\ref{fig:alignment}.
Figures~\ref{fig:dynamics_in_variance_limited_regime_mnist},~\ref{fig:internal_scaling_mechanisms},~\ref{fig:scaling_phases},~\ref{fig:network_response},~\ref{fig:effective_rank}, and ~\ref{fig:alignment},
we consider the following experimental setups aligning 
with the setups from~\citet{bahriExplainingNeuralScaling2024b}.

\textbf{Model Scaling}:
Each dataset was trained with a three-and four-layer dense NN of varying widths 
using LeCun initialization~\citep{lecunEfficientBackProp1998}, ReLU activations, and
Adam optimization~\citep{kingmaAdamMethodStochastic2017b} with a momentum of 0.9 and a 
learning rate small enough to ensure a minimum test loss is reached between 400 and 1000 epochs.
The training was carried out on 100 randomly chosen samples with full-batch gradient 
descent, cross-entropy loss and the neural tangent kernel was computed on 128 test samples.
The ensembles are performed over 20 draws of the network initialization. 

The results of the three-layer model are presented in Section~\ref{sec:scaling_mechanisms}
and~\ref{sec:feature_kernel_transition}, and the four-layer model results are shown in 
Figure~\ref{fig:scaling_phases} to help understand the transition from feature to kernel-like behavior.
% The full results of the four-layer model are discussed in Appendix~\ref{sec:app_results_scaling_experiments}.

\textbf{Data Scaling}:
Each dataset was trained with a four-layer dense NN of 8 units each
using LeCun initialization~\citep{lecunEfficientBackProp1998}, ReLU activations, and
Adam optimization~\citep{kingmaAdamMethodStochastic2017b} with a momentum of 0.9 and a 
learning rate small enough to 
ensure a minimum test loss is reached between 400 and 1000 epochs.
The training was carried out using full-batch gradient descent, cross-entropy loss and
the neural tangent kernel was computed on 128 test samples.
The ensembles are performed over 20 draws of the training data.

\subsection{Noise Scaling} \label{sec:noise_scaling}
In this study, networks are trained on the CIFAR-4 (consisting of the classes "birds", "cats", "airplanes", and "automobiles"). 
Additionally, we reduce the information of the training dataset to scale into the data bottlenecked regime.
The training data of 2000 samples is augmented in the following way:
For each class, we replace the training samples by noisy versions of other samples from the same class, 
using a Gaussian noise with a standard deviation of 0.01. 
Both, the replaced and the replacing samples are chosen randomly, such that $n$ images
equally replace $m$ images of the same class, with $n + m$ equal to the total number of samples 
in one class of the training data. 
This way, we control the amount of class-internal features in the training data\footnote{
    Note that we explicitly replace samples from each class rather than removing samples. 
    This way, we can ensure the dynamics take place on the same time scale, 
    enabling to compare runs in terms of epochs. 
}.
The test dataset on which we measure the NTK trace and effective rank is kept unchanged. \\
The dense network consists of 3 hidden layers with 256 units each, ReLU activations, and
Xavier-uniform initialization~\citep{glorotUnderstandingDifficultyTraining2010} and 
the convolutional network consists of 3 convolutional layers with 32 filters each, followed 
by a fully connected layer with 64 units, all using ReLU activations and He-uniform initialization~\citep{heDelvingDeepRectifiers2015}.
The models were trained using SGD with a learning rate of 0.001 and a momentum of 0.9 for 1000 epochs using a batch size of 128. 
The training was carried out using cross-entropy loss and neural tangent kernel measures were ensembled over three draws 
of 400 test samples.

\section{Supplemental Results} \label{sec:supplemental_results}

\subsection{Scaling Experiments} \label{sec:app_results_scaling_experiments}
In the following, we present the dynamics results for all datasets utilized in the scaling experiments
described in Section~\ref{sec:scaling_mechanisms} and presented in Figure~\ref{fig:internal_scaling_mechanisms}.
The exact setup of the experiments is described in \cref{sec:variance_scaling}.
% This includes the loss scaling behavior for all investigated datasets and architectures, 
% as well as the NTK trace and effective rank dynamics for a four-layer multi-layer perceptron (MLP).

\begin{figure*}[h]
    \begin{center}
    \includegraphics[width=1\linewidth]{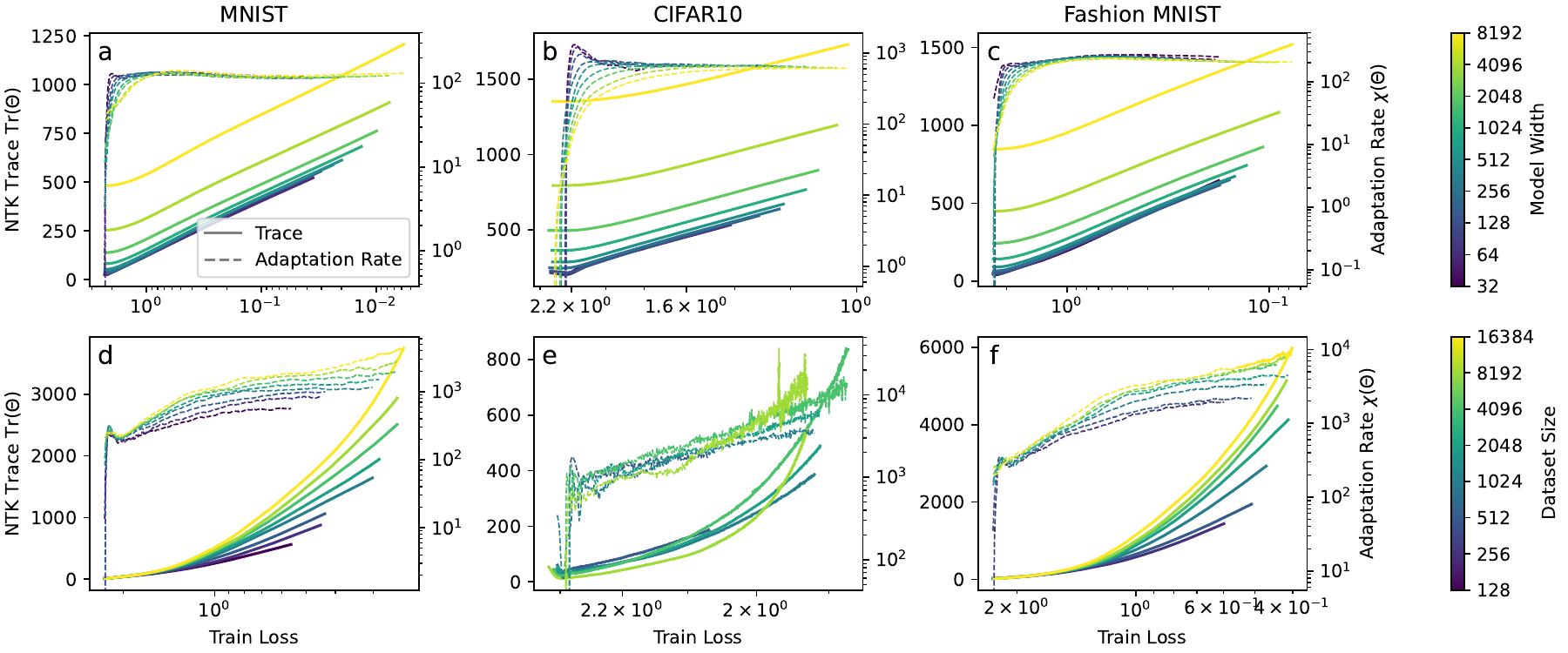}
    \end{center}
    \caption{
        Dynamics of the NTK trace $\text{Tr} (\bm{\Theta})$ and adaptation rate $\chi(\bm{\Theta})$ as a function of training loss on
        standard vision datasets. 
        The figure shows learning curves for various model widths (a - c) and dataset sizes (d - f), spanning multiple scales.
        Each curve tracks the evolution from the initial model state to the point of minimum test loss. 
        Results are averaged over 20 ensembles.
        }
    \label{fig:network_response}
% \end{figure*}
% % \paragraph{Effective Rank}
% \begin{figure*}[h]
    \begin{center}
    \includegraphics[width=\linewidth]{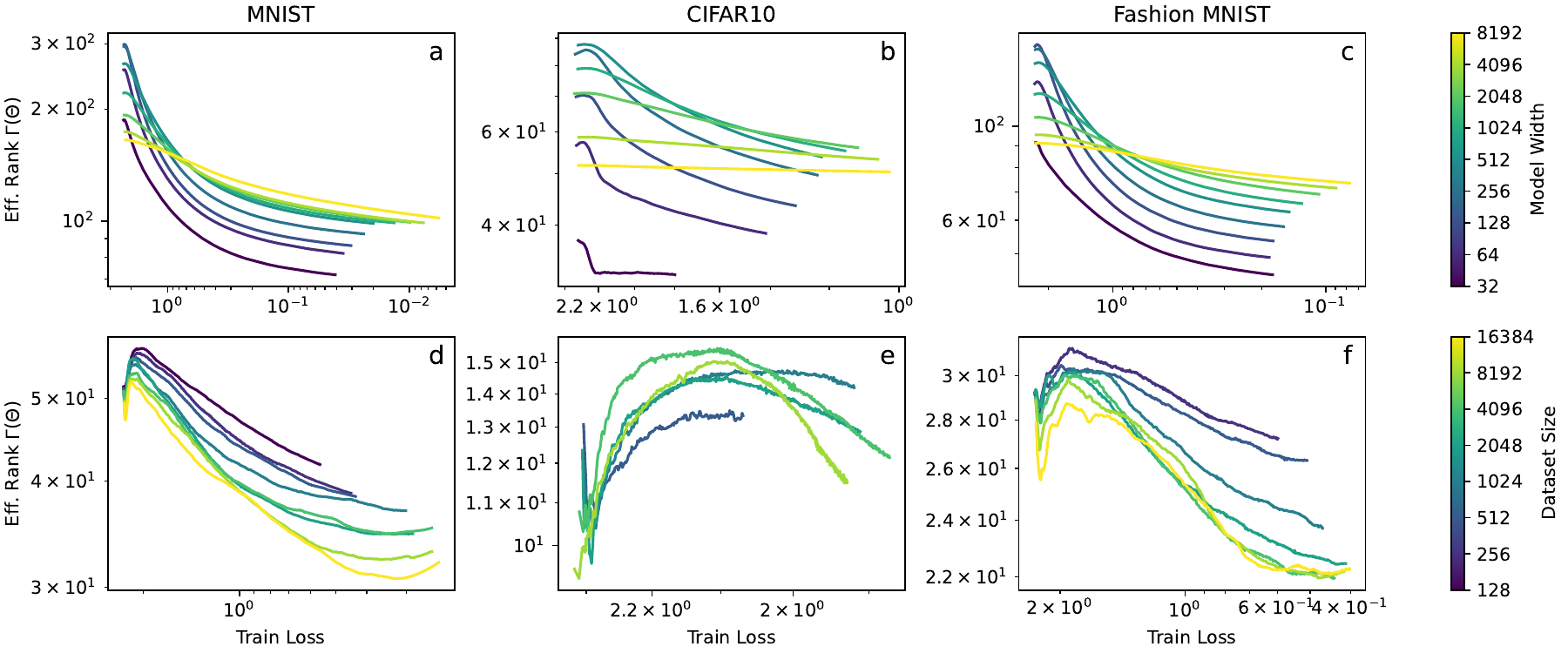}
    \end{center}
    \caption{
        Dynamics of the Effective Rank $\Gamma  (\bm{\Theta})$ of the NTK as a function of training loss on
        standard vision datasets. 
        The figure shows learning curves for various model widths (a - c) and dataset sizes (d - f) spanning multiple scales.
        Each curve tracks the evolution from the initial model state to the point of minimum test loss. 
        Results are averaged over 20 ensembles.
    }
    \label{fig:effective_rank}
\end{figure*}

In Figure~\ref{fig:network_response}, we present the NTK trace and adaptation rate dynamics for scaling model and data sizes.
In Figure~\ref{fig:effective_rank}, we present the effective rank dynamics for scaling model and data sizes. \\
The trace and adaptation rate results show that the response of model and data scaling behaves as described in Section~\ref{sec:dynamics_in_variance_limited_regime}, 
allowing us to utilize the initial trace and the adaptation rate at minimum test loss as key observables to quantify the dynamics. 
The effective rank results depict the same sorting behavior as described in Section~\ref{sec:dynamics_in_variance_limited_regime}. 

In addition to the analysis presented in the main body of this work, in the following, we want to point out an additional observation.\\
The effective rank dynamics from \cref{fig:effective_rank} depict surprisingly similar behavior across all datasets:
Looking at the dynamics trained on MNIST, then Fashion-MNIST, and finally CIFAR-10, one could argue that the dynamics of the effective rank are similar but take place on different "time" scales.
For example, for scaling dataset size, the small peak in the effective rank dynamics of the MNIST dataset is present in the Fashion-MNIST dataset as well, but its extension is much shorter.
Training on CIFAR-10, the entire training is spent in the peak region. 
A similar observation can be made for the dynamics in the model scaling regime. 
This suggests that the stages a model evolves through might be the same. 
The difference in the observed model dynamics might lie in the "time" scale of this evolution, which seems to be affected by the complexity of the data:
The more complex the data, the slower the process. 
% a main difference in the observed model dynamics might lie in the "time" scale at which a model moves through the phases of learning. 

\subsection{Noise Scaling} \label{sec:noise_scaling_results}
\begin{figure}[h]
    \begin{center}
    \includegraphics[width=1\linewidth]{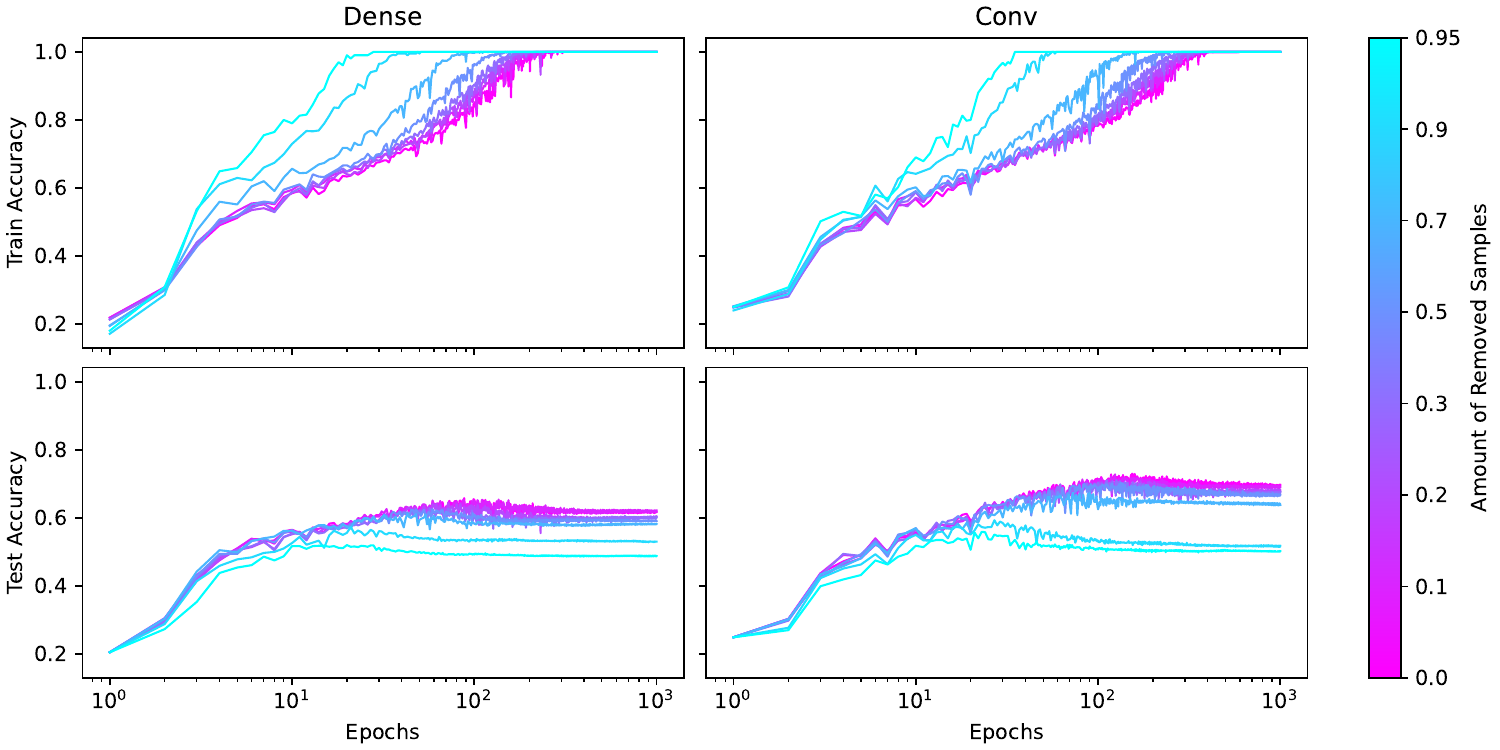}
    \end{center}
    \caption{
            Train and test accuracies of dense and convolutional NNs trained on CIFAR-4.
            We create datasets containing different amounts of features by removing varying numbers of training samples and replacing them with noisy copies of the remaining samples.
            A detailed description of the experiment can be found in \cref{sec:noise_scaling}.
        }
    \label{fig:controlled_dynamics_acc}
\end{figure}

In the following, we present additional results of the noise scaling experiment shown in Section~\ref{sec:effective_rank_feature_learning_app} and Appendix~\ref{sec:noise_scaling}.
The performance of dense and convolutional NNs trained on CIFAR-4 with varying amount of removed samples in the training data is presented in
Figure~\ref{fig:controlled_dynamics_acc}.
The results show that all networks are trained to converge in train loss, and that performance decreases as the amount of removed data increases.

\subsection{Relation to Alignment} \label{sec:alignment_theories} %\looseness=-1 

\begin{figure}[h]
    \begin{center}
    \includegraphics[width=1\linewidth]{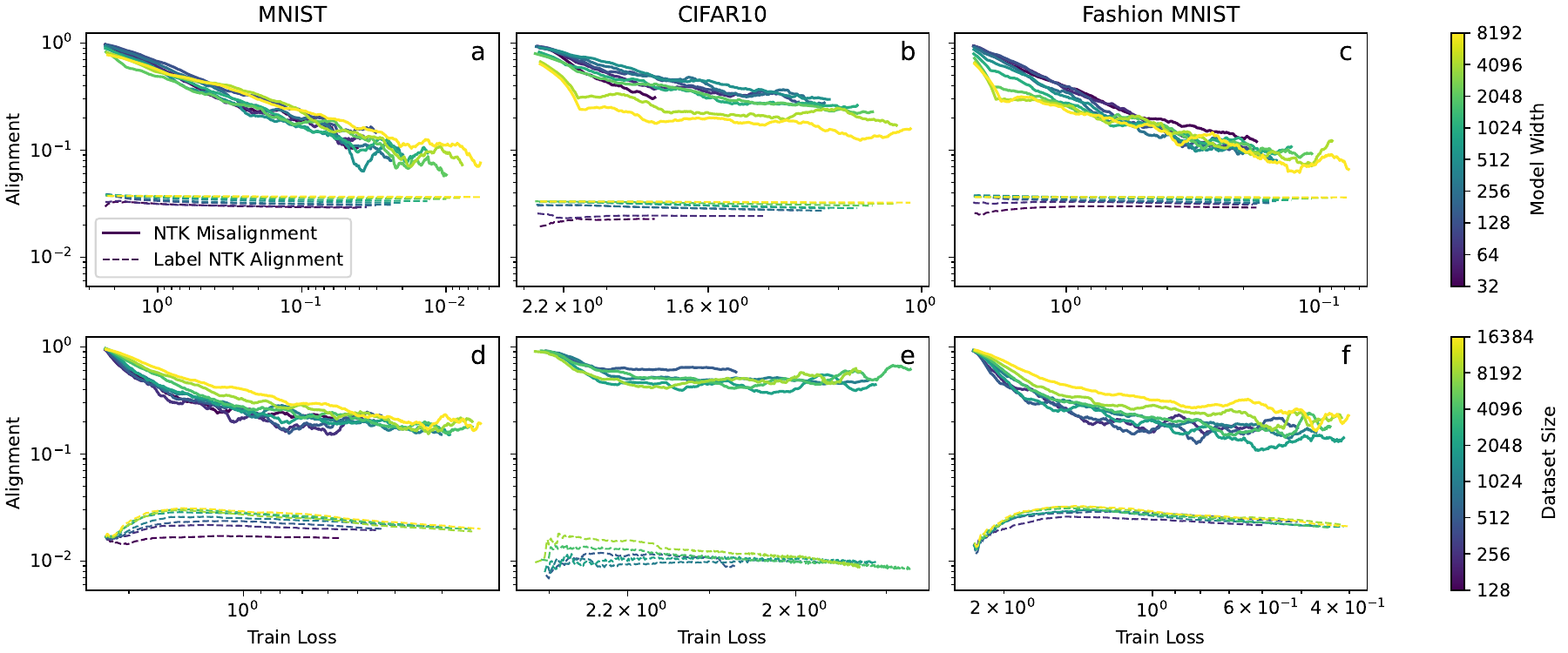}
    \end{center}
    \caption{
        Dynamics of the Label NTK alignment and NTK misalignment as a function of training loss on standard vision datasets. 
        The figure shows learning curves for various model widths (a - c) and dataset sizes (d - f) spanning multiple scales.
        Each curve tracks the evolution from the initial model state to the point of minimum test loss. 
        Results are averaged over 20 ensembles.
        }
    \label{fig:alignment}
\end{figure}

% Relation to other works
% Why can we claim to have new mechanisms?
A line of work has studied network dynamics via the alignment of the 
NTK to the task at hand~\citep{baratinImplicitRegularizationNeural2021, shanTheoryNeuralTangent2022, atanasovNeuralNetworksKernel2021,ortiz-jimenezWhatCanLinearized2021}.
To integrate our results into the notion of kernel alignment, this section presents alignment measures during the scaling study. 
Figure~\ref{fig:alignment} shows the label-alignment of the NTK computed via 
\begin{equation}
    \text{Label NTK Alignment} = \frac{ \langle \bm{\Theta} , Y Y^\top \rangle_F}{\|Y Y^\top\|_F \|\bm{\Theta}\|_F},
\end{equation}
where $Y$ is the label matrix, $\| \cdot \|_F$ denotes the Frobenius norm and $\langle \cdot , \cdot \rangle_F$ the Frobenius inner product.
In addition, we compute the misalignment of the top-n eigenvectors of the NTK 
after updating the network parameters,
\begin{equation}
    \text{NTK Misalignment} = 1 - \frac{\langle \bm{Q}_\Gamma^\top , \bm{Q}_\Gamma^\prime \rangle_F}{\| \bm{Q}_\Gamma \|_F \| \bm{Q}_\Gamma^\prime \|_F}.
\end{equation}
Here, $\bm{Q}$ and $\bm{Q}^\prime$ are the eigenvector matrices of the NTK before, 
and after updating the network parameters, respectively.
We choose the number of dominant eigenvectors to be given via the effective rank, 
indicating by denoting the top-$\Gamma$ eigenvectors with an index\footnote{
    Selecting the top-$\Gamma$ eigenvectors instead of all eigenvectors is done to 
    reduce the noise introduced by the eigenvectors with small eigenvalues.
}.
We compare this quantity to the amount of directional alignment in the NTK, 
as each rotation of NTK eigenvectors could change the directional alignment of the NTK 
at the same rate. 

% Observations
From the results in Figure~\ref{fig:alignment}, we can observe that eigenvector rotation
starts close to the maximum value and decreases over time.
The label alignment, however, only barely changes. 
This means that from the dynamic reorientation, the label alignment only slightly benefits, 
suggesting that effects other than label alignment are dominating the investigated 
scaling regimes.

% Interpretation
This leads us to the conclusion that the effects we observed via the effective rank and trace do not seem to be related to label alignment.
Consequently, our results appear to be \textit{additional} effects besides alignment-based theories. 

%%%%%%%%%%%%%%%%%%%%%%%%%%%%%%%%%%%%%%%%%%%%%%%%%%%%%%%%%%%%%%%%%%%%%%%%%%%%%%%
%%%%%%%%%%%%%%%%%%%%%%%%%%%%%%%%%%%%%%%%%%%%%%%%%%%%%%%%%%%%%%%%%%%%%%%%%%%%%%%

\end{document}